\documentclass[letterpaper]{article} % DO NOT CHANGE THIS
\usepackage{aaai25}  % DO NOT CHANGE THIS
\usepackage{times}  % DO NOT CHANGE THIS
\usepackage{helvet}  % DO NOT CHANGE THIS
\usepackage{courier}  % DO NOT CHANGE THIS
\usepackage[hyphens]{url}  % DO NOT CHANGE THIS
\usepackage{graphicx} % DO NOT CHANGE THIS
\usepackage{natbib}  % DO NOT CHANGE THIS AND DO NOT ADD ANY OPTIONS TO IT
\usepackage{caption} % DO NOT CHANGE THIS AND DO NOT ADD ANY OPTIONS TO IT
\usepackage{algorithm}
\usepackage{algorithmic}
\nocopyright
\usepackage{newfloat}
\usepackage{listings}
\usepackage{amsmath} % Added for math fonts
\DeclareCaptionStyle{ruled}{labelfont=normalfont,labelsep=colon,strut=off} % DO NOT CHANGE THIS
\floatstyle{ruled}
\newfloat{listing}{tb}{lst}{}
\floatname{listing}{Listing}
\title{Multilinguality in LLM-Designed Reward Functions for Restless Bandits: Effects on Task Performance and Fairness}

\author{
    Ambreesh Parthasarathy\textsuperscript{\rm 1},
    Chandrasekar Subramanian\textsuperscript{\rm 1},
    Ganesh Senrayan\textsuperscript{\rm 2},
    Shreyash Adappanavar\textsuperscript{\rm 2},
    Aparna Taneja\textsuperscript{\rm 3},
    Balaraman Ravindran\textsuperscript{\rm 1 \rm 2},
    Milind Tambe\textsuperscript{\rm 3}
}

\affiliations{
    \textsuperscript{\rm 1}Centre for Responsible AI, Indian Institute of Technology Madras\\
    \textsuperscript{\rm 2}Indian Institute of Technology Madras\\
    \textsuperscript{\rm 3}Google Deepmind\\
    sekarnet@gmail.com
}

\usepackage{bibentry}
\begin{document}
% \title{Fair-RMABs}
\maketitle

\begin{abstract}
Restless Multi-Armed Bandits (RMABs) have been successfully applied to resource allocation problems in a variety of settings, including public health. With the rapid development of powerful large language models (LLMs), they are increasingly used to design reward functions to better match human preferences. Recent work has shown that LLMs can be used to tailor automated allocation decisions to community needs using language prompts. However, this has been studied primarily for English prompts and with a focus on task performance only. This can be an issue since grassroots workers, especially in developing countries like India, prefer to work in local languages, some of which are low-resource. Further, given the nature of the problem, biases along population groups unintended by the user are also undesirable. In this work, we study the effects on both task performance and fairness when the DLM algorithm, a recent work on using LLMs to design reward functions for RMABs, is prompted with non-English language commands.  Specifically, we run the model on a synthetic environment for various prompts translated into multiple languages. The prompts themselves vary in complexity. Our results show that the LLM-proposed reward functions are significantly better when prompted in English compared to other languages. We also find that the exact phrasing of the prompt impacts task performance. Further, as prompt complexity increases, performance worsens for all languages; however, it is more robust with English prompts than with lower-resource languages. On the fairness side, we find that low-resource languages and more complex prompts are both highly likely to create unfairness along unintended dimensions.
\end{abstract}


\section{Introduction}
Reward functions are crucial to the generation of optimal policies for sequential decision-making via reinforcement learning. Previous works \cite{behari2024decisionlanguagemodeldlmdynamic, verma2024balancingactprioritizationstrategies} have shown that LLMs are quite useful when it comes to designing reward functions that are aligned with human preferences.  However, these existing methods only study this problem for human commands that are in English. Additionally, they only focus on task performance (i.e., how well the given commands are executed). 

However, as seen in both papers, a key domain of application is public health. Specifically, the DLM algorithm \cite{behari2024decisionlanguagemodeldlmdynamic}  has been used to allocate health worker calls to underprivileged socioeconomic groups. This can be used to empower grassroots health workers to specify commands based on community needs, and the algorithm would automatically tailor allocation decisions based on that. However, two questions remained unanswered: (1) What happens if the prompts are not in English? (2) Is there an impact on the fairness of the allocations? We are not aware of any work that studies these questions for the DLM algorithm.

This is especially important to look at in developing countries like India where these workers might prefer to work in local languages, some of which are low-resource languages. Furthermore, given the nature of the problem, unintended biases along population groups are also undesirable.  

This paper studies the effects on both task performance and fairness when the DLM algorithm \cite{behari2024decisionlanguagemodeldlmdynamic} is prompted in non-English languages. 
\paragraph{Related work} There has been a lot of recent work in using LLMs to design reward functions for reinforcement learning settings. Works by 
\cite{sun2024largelanguagemodeldrivenreward, kwon2023rewarddesignlanguagemodels, ma2024eureka, pmlr-v229-yu23a, Cao_2024, behari2024decisionlanguagemodeldlmdynamic, verma2024balancingactprioritizationstrategies},  
% \cite{sun2024largelanguagemodeldrivenreward}, \cite{kwon2023rewarddesignlanguagemodels}, \cite{ma2024eurekahumanlevelrewarddesign}, \cite{yu2023languagerewardsroboticskill} and \cite{Cao_2024} 
among others, have shown that using LLMs to design reward functions allows the reinforcement learning modules to be more flexible and helps them ground their outcomes in natural language. % To that end \cite{behari2024decisionlanguagemodeldlmdynamic} and \cite{zhao2024banditwhisperercommunicationlearning} show that LLMs hold great promise in the case of Restless Multi-armed Bandits as well. 
% Given that RMABs can be used for resource allocation tasks at grassroots level and for domains where resource allocation is a critical task (rewrite), it is important we analyze and understand if there is any difference in outcomes for users engaging in different languages.\\

There has also been work that has looked at the multilinguality of LLMs from the perspectives of improving task performance  \cite{lai-etal-2024-llms} and bias mitigation \cite{demidova-etal-2024-john}.

However, to our knowledge, ours is the first work that studies the impact on task performance and fairness when prompts to LLMs (which are used to design reward functions) are in non-English languages. 
\section{Experiment Setup}
Our experiments are intended to answer the following research questions:
\begin{enumerate}
    \item How does language affect reward function proposals? We study this in Section \ref{sec:analysis-1-acceptable-reward}.
    \item How do non-English prompts affect task performance?  We study this in Section \ref{sec:analysis-2-task-perf}.
    \item How do non-English prompts affect fairness? We study this in Section \ref{sec:analysis-3-fairness}.
%%%%
    % \item Does Language have an effect on the performance of the DLM Algorithm?
    % \item Do Language and Prompt Complexity have any effects on the other feature dimensions thereby leading to unintended consequences?
\end{enumerate}

We take the DLM algorithm \cite{behari2024decisionlanguagemodeldlmdynamic} and run a variety of prompts in different languages. The specific languages we consider are English, Hindi, Tamil, and Tulu (a low-resource language). We then look at the resource allocation to various subgroups (defined by different values of the  features) and perform various analyses on task performance and fairness. Note that, for our analyses, we consider the allocations at the end of the evolutionary search of the DLM algorithm. 
Following \cite{verma2024balancingactprioritizationstrategies}, we use Gemini 1.0 Pro \cite{team2023gemini} as the LLM. We also use a solution based on the Whittle Index \cite{whittle1988restless} for the reinforcement learning portion, similar to \cite{verma2024balancingactprioritizationstrategies}.

\subsection{Environment}
We consider 6 features taken from \cite{behari2024decisionlanguagemodeldlmdynamic} for this paper. However, we use a synthetic environment to perform our analyses instead. We describe the environment below. We generate features based on the schema we have described below. Given the difficulties in acquiring real-world data, we instantiate a synthetic dataset, where the features and effects are reflective of real-world settings. We derive our transition probabilities from these features. 

% \textit{Note}: While the features are based off of phenomenon observed in the real world, it is not imperative for it to be 1:1 replica, as different populations might have different dynamics to these features.

\subsubsection{Initial Feature Distribution} We assume that the initial feature distribution is generated from the following graphical model.\footnote{See \cite{Pearl_2009, koller_pgm} for a primer on graphical models.}
The structural relationships are described as follows:

\begin{align*}
    \textit{Age} &\to \textit{Income} \\
    \textit{Education\_Level} &\to \textit{Income} \\
    \textit{Income} &\to \textit{Phone\_Ownership} \\
    \textit{Phone\_Ownership} &\to \textit{Times\_To\_Be\_Called} \\
    % \text{Language Spoken} &\text{ (independent)}
\end{align*}
\textit{Language\_Spoken} is independent of all other variables. 

We also choose a constant $\alpha$ that denotes the strength of the relationships between the variables. We assume that the structural equations are as follows:

\begin{equation*}
    \textit{Age} \leftarrow \text{Unif}[0, 1]
\end{equation*}

\begin{equation*}
    \textit{Education\_Level} \leftarrow  \text{Unif}[0, 1]
\end{equation*}

\begin{equation*}
    \textit{Language\_Spoken} \leftarrow  \text{Unif}[0, 1]
\end{equation*}

\begin{multline*}
    \textit{Income} \leftarrow  \text{Clip} \Big\{ 
    \alpha \cdot \frac{\textit{Age} + \textit{Education}}{2} + \\
    (1 - \alpha) \cdot \text{Unif}[0, 1] 
    \Big\}
\end{multline*}

\begin{multline*}
    \textit{Phone\_Ownership} \leftarrow  \text{Clip} \Big\{ 
    \alpha \cdot (1 - \textit{Income}) + \\
    (1 - \alpha) \cdot \text{Unif}[0, 1] 
    \Big\}
\end{multline*}

\begin{multline*}
    \textit{Times\_To\_Be\_Called} \leftarrow  \text{Clip} \Big\{ 
    \alpha \cdot \textit{Phone\_Ownership} + \\
    (1 - \alpha) \cdot \text{Unif}[0, 1] 
    \Big\}
\end{multline*}

The computed quantities for each feature are then bucketed into its possible values. Refer to Table \ref{tab:feature-buckets} in Appendix \ref{app:variable-values} for the values of each feature. 

% [TBD - paragraph justifying both the graphical model and the structural equations]\\

Similar to \cite{verma2024balancingactprioritizationstrategies}, we have assumed that the variables follow a uniform distribution as we do not have any information on the prior distribution of these variables. However, we intend to use more realistic distributions based on real-world data in future work. We have chosen the above structural equations for the following reasons. \textit{Age} affects \textit{Income} because older people are more likely to have more revenue streams. \textit{Education} affects \textit{Income} because more education allows better income opportunities. \textit{Phone\_Ownership} affects \textit{Times\_To\_Be\_Called} because, the more likely the phone is owned by a family, the more likely they would like to be called at a time when they have assured privacy. \textit{Income} affects \textit{Phone\_Ownership} as more income could result in less likelihood of group ownership of the phone. In our structural equations, we assume \textit{Language\_Spoken} to be independent of the rest.

We consider two possible values for $\alpha$: 0.2 and 0.8. The former is intended to simulate a low conditional correlation between nodes of our graphical model, and the latter is intended to simulate a higher conditional correlation.

\subsubsection{Transition Probabilities}
Transition probabilities are computed in the same way as in \cite{verma2024balancingactprioritizationstrategies}. We use the “weight vector parameters” of the features inspired by Table 2 in the paper. There are two states, 0 and 1. The good state (state 1) is when beneficiaries listen to an automated message for more than 30 seconds, and the bad state (state 0) is when the participant does not engage.

The key idea is that the values of the features and the weights are used to calculate a value $\delta$, which is the difference between the active transition probabilities and the passive transition probabilities that we observe in a Restless MAB \cite{10.1109/TIT.2012.2230215}. By expressing this $\delta$ as a function of the feature values and dataset weights, we create a dependency for the active transition probabilities on the features.
Table \ref{tab:weight-vector-params} provides the weights for each of the features.

% [TBD - Mini-explanation of the $\delta$ and how the above will convert to TPs]\\
% 1 or 2 lines general idea (lp) (TBD)

\begin{table}[h]
    \centering
    \renewcommand{\arraystretch}{1.25}
    \begin{tabular}{|c|c|} \hline
    \textbf{Feature} & \textbf{Weight}\\ \hline
         \textit{Age}& 0.8\\ \hline 
         \textit{Income}& 1.5\\ \hline 
         \textit{Language\_Spoken}& -0.3\\ \hline 
         \textit{Education\_Level}& 1.5\\ \hline 
         \textit{Phone\_Ownership}& -1.5\\ \hline 
         \textit{Times\_To\_Be\_Called}& 0.3\\ \hline
    \end{tabular}
    \caption{Parameters for the weight vector}
    \label{tab:weight-vector-params}
\end{table}
 The magnitudes of the weights are inspired by \cite{verma2024balancingactprioritizationstrategies}, and their directions, corresponding to their signs, are obtained as follows:
\begin{enumerate}
    \item Older mothers are more likely to listen to the calls and, hence, more likely to be converted to a good state. Thus, \textit{Age} warrants a positive value.
    
    \item \textit{Income} has a larger positive weight since higher-income people are more likely to engage with the calls.

    \item The women are more likely to engage with the call if they own the phone. Thus, a large negative value is implied for \textit{Phone\_Ownership}

    \item More educated women are more likely to be converted; thus, \textit{Education\_Level} has a larger positive value.
    
    \item \textit{Language\_Spoken} does not influence the likelihood of conversion; thus, it can take a small positive/negative value.
    
    \item Although not a strong factor, women who pick up the call late in the evening or at night are more likely to be converted. Thus, \textit{Times\_To\_Be\_Called} has a small positive value.

    \end{enumerate}
Note that some numbers are negative; this means that the lower-indexed buckets have a higher probability of moving to a good state. 
     % The directions tell which buckets are more likely to convert; positive meaning the end-buckets and negative meaning the start buckets. 

\subsubsection{Rewards} 
The reward is a function of the resultant state (the state that the arm reaches after the action has been completed) and the features of the arm. Refer to Table \ref{tab:full-prompt} in the Appendix \ref{app:full-prompt} to see some examples of reward functions.

\subsection{Prompts}
\begin{table}[h!]
    \centering
    \renewcommand{\arraystretch}{1.5} % Adjust row height (padding)
    \begin{tabular}{|p{0.45cm}|p{4cm}|p{2.8cm}|}
    \hline
    \textbf{No.} & \textbf{Prompt} & \textbf{Features} \\ \hline
    1 & While still prioritizing all, slightly prioritize those who have a low value of age. & \textit{Age} \\ \hline
    2 & While still prioritizing all, slightly focus on middle-aged mothers earning less than 15000. & \textit{Age}, \textit{Income} \\ \hline
    3 & While prioritizing all, slightly focus more on mothers who are busy during mornings and afternoons. & \textit{Times\_To\_Be\_Called} \\ \hline
    4 & While still prioritizing all, slightly advantage those who prefer being called between 12:30 pm - 3:30 pm and the woman is more likely to take the call. & \textit{Times\_To\_Be\_Called}, \textit{Phone\_Ownership} \\ \hline
    5 & While prioritizing all, slightly focus on South Indian mothers with a poor educational background. & \textit{Language\_Spoken}, \textit{Education\_Level} \\ \hline
    6 & While still prioritizing all, place a slightly stronger emphasis on older women with lower incomes who are Kannadiga. & \textit{Age}, \textit{Income}, \textit{Language\_Spoken} \\ \hline
    7 & While prioritizing all, slightly focus on young mothers. & Rephrasing Prompt 1 \\ \hline
    8 & While still prioritizing all, focus on mothers in their 30s or 40s, and who also earn less than 500 a day. & Rephrasing Prompt 2 \\ \hline
    \end{tabular}
    \caption{Prompts used}
    \label{tab:prompts}
\end{table}
We use a total of 8 prompts (also called `goal prompts') for our analysis. Prompts 1 through 6 are increasing in complexity, while Prompts 7 and 8 are rephrased versions of Prompts 1 and 2, respectively. These will be used to analyze the effects of prompt complexity and prompt rephrasing. As we will see, outcomes indeed vary along both dimensions and in Section \ref{sec:future-work}, we propose one direction that can be explored in the future to mitigate the issues with respect to phrasing. The full set of prompts is given in Table \ref{tab:prompts}. To view the full prompt given to the LLM, refer to Table \ref{tab:full-prompt} in Appendix \ref{app:full-prompt}.

\subsection{Fairness Metrics}
%\subsubsection{Kullback-Leibler (KL) Divergence}

%The Kullback-Leibler (KL) Divergence is a measure of the difference between two probability distributions. In the context of fairness, it can be used to quantify how much the allocation distribution for a given prompt deviates from the baseline distribution. 

%Here, we calculate the KL Divergence between the allocation distribution for the base prompt \textit{"Prioritize everyone equally"} (denoted as \(P\)) and the allocation distribution for the intended prompt (denoted as \(Q\)). The KL Divergence is given by:

%\begin{equation}
 %   D_{KL}(P \| Q) = \sum_{i=1}^{k} P(X_i) \log \frac{P(X_i)}{Q(X_i)},
%\end{equation}

%where:
%\begin{itemize}
%   \item \(P(X_i)\): Probability of allocating resources to group \(X_i\) under the base prompt.
%    \item \(Q(X_i)\): Probability of allocating resources to group \(X_i\) under the intended prompt.
%    \item \(k\): Number of demographic groups or categories.
%\end{itemize}

%This divergence value represents the information gain or the additional effort required to move from the baseline distribution \(P\) to the intended distribution \(Q\). A lower value of \(D_{KL}\) indicates that the intended allocation distribution closely aligns with the baseline, implying minimal deviation from the base prompt.

%\begin{equation}
%D_{KL}(P \| Q) = \sum_{x \in X} P(x) \log \frac{P(x)}{Q(x)}   
%\end{equation}
\subsubsection{Demographic Parity}

Demographic Parity \cite{NIPS2017_a486cd07} ensures that the outcome of a model is independent of a specific protected attribute (e.g., gender, race, or age). It requires that individuals from different demographic groups  have the same probability of being allocated.

Consider the feature \textit{Age}, which can take values in discrete intervals, such as \(10\text{-}20\), \(21\text{-}30\), \(31\text{-}40\), \(41\text{-}50\), and \(51\text{-}60\). The probability of an arm falling in one of these buckets (say, $x$) receiving an allocation (which is given by $Y=1$) is
%Demographic Parity is defined as the probability of being allocated a positive outcome conditioned on the feature. For instance, Demographic Parity for a specific age group \(x\) is given by:

\begin{equation}
    P(Y = 1 \mid \text{Age} = x) = \frac{P(\text{Age} = x \mid Y = 1) \cdot P(Y = 1)}{P(\text{Age} = x)},
\end{equation}

where \(Y \in \{0, 1\}\) is the random variable that is 1 if a particular arm of the Restless MAB has received an allocation. Here, the age group can take values \(x_1, x_2, \ldots, x_k\), corresponding to the discrete intervals of the feature. 

To evaluate the fairness of the system across different demographic groups, we compute the variance in the allocation probabilities for different subgroups. This metric, referred to as \textit{DP\_variance}, is defined as:

% \begin{equation}
%     DP\_variance = \frac{1}{k} \sum_{i=1}^{k} \left(P(Y = 1 \mid \text{Age} = x_i) - \overline{P(Y = 1)}\right)^2,
% \end{equation}

\begin{multline}
    DP\_variance = \\ \frac{1}{k} \sum_{i=1}^{k} \left(P(Y = 1 \mid \text{Age} = x_i) - \overline{P(Y = 1)}\right)^2
\end{multline}

where \(k\) is the total number of groups (e.g., age intervals), and \(\overline{P(Y = 1)}\) is the mean probability of receiving an allocation, calculated as:

\begin{equation}
    \overline{P(Y = 1)} = \frac{1}{k} \sum_{i=1}^{k} P(Y = 1 \mid \text{Age} = x_i).
\end{equation}

We consider a lower \textit{DP\_variance} to indicate that the model is closer to satisfying Demographic Parity. This metric can serve as an indicator of fairness, with zero variance implying perfect Demographic Parity.

\section{Analysis 1: Proposing Acceptable Reward Functions}
\label{sec:analysis-1-acceptable-reward}
We define an acceptable reward function as a reward proposal by the LLM that contains all the relevant features and no spurious features. In the case where the goal prompt contains multiple features, an acceptable reward function must select the right features for all the conditions. Note that we consider the final reward functions proposed at the end of the evolutionary search of the DLM algorithm.

We measure the rate of acceptable reward functions for the languages English, Hindi, Tamil and Tulu and report it as $Mean \pm StdError$. Note that the error reported is one standard error.

\subsection{Result 1: English Outperforms in Proposing Acceptable Reward Functions}

From Table \ref{tab:acceptable-prompt-rates}, we observe that English has a higher rate of acceptable reward functions proposed by the LLM. We also observe from the allocation plots in Appendix \ref{app:alloc-plots} that this higher rate of acceptable reward functions proposed is directly correlated with task performance. This could be due to the LLM being provided with a better set of allocations for reflection, thereby improving the overall performance.

\begin{table*}[h!]
\centering
\renewcommand{\arraystretch}{1.25}
\small % Reduces the font size for the table
\begin{tabular}{|c|p{2.2cm}|p{2.2cm}|p{2.2cm}|p{2.2cm}|p{2.2cm}|p{2.2cm}|}
\hline
\textbf{Language} & \textbf{Prompt 1} & \textbf{Prompt 2} & \textbf{Prompt 3} & \textbf{Prompt 4} & \textbf{Prompt 5} & \textbf{Prompt 6} \\ \hline
\textbf{English} & $0.65 \pm 0.15$ & $0.033 \pm 0.017$ & $0.033 \pm 0.017$ & $0.7 \pm 0.25$ & $0.10 \pm 0.05$ & $0.083 \pm 0.017$ \\ \hline
\textbf{Hindi}   & $0.6 \pm 0.15$ & $0.033 \pm 0.017$ & $0.033 \pm 0.017$ & $0.25 \pm 0.05$ & $0.10 \pm 0.05$ & $0.033 \pm 0.017$ \\ \hline
\textbf{Tamil}   & $0.433 \pm 0.15$ & $0.083 \pm 0.017$ & $0.033 \pm 0.017$ & $0.45 \pm 0.05$ & $0.10 \pm 0.083$ & $0.033 \pm 0.017$ \\ \hline
\textbf{Tulu}    & $0.45 \pm 0.15$ & $0.0 \pm 0.0$ & $0.0 \pm 0.0$ & $0.35 \pm 0.25$ & $0.10 \pm 0.083$ & $0.00 \pm 0.00$ \\ \hline
\end{tabular}
\caption{Acceptable prompt rates with 1 standard error for each language and prompt}
\label{tab:acceptable-prompt-rates}
\end{table*}

\section{Analysis 2: Task Performance Under Multilingual Prompts}
\label{sec:analysis-2-task-perf}
% TBD. \\
We measure task performance (i.e., how well the allocations align with the prompt) for multilingual prompts along two dimensions.
\begin{enumerate}
    \item \textbf{Phrasing}: We plot the allocation values for relevant features across different languages for two prompts that are semantically identical but phrased differently.
    \item \textbf{Prompt Complexity}: We plot the rate of acceptable allocations across various runs. We delve deeper into what an acceptable allocation is in later sections.
\end{enumerate}

All results are averaged over 20 independent runs, and error bars are $\pm$ 1 standard error from the sample mean.

\subsection{Result 2: Phrasing Makes a Difference in Performance}.
\label{res:result2}

\begin{figure}[h!]
    \centering
    \includegraphics[width=0.5\textwidth]{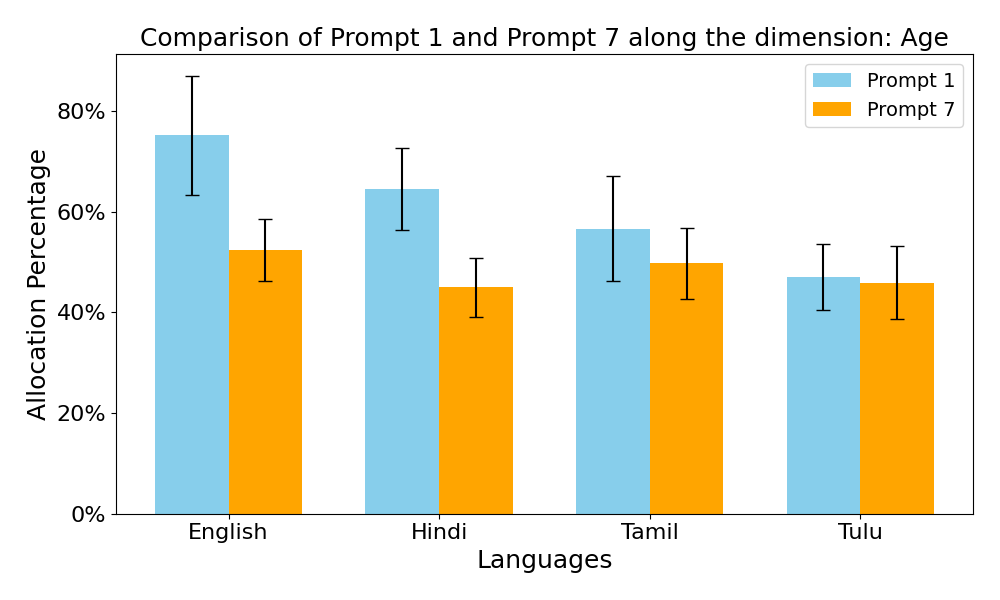}
    \caption{Plot showcasing the allocation percentages summed for the age values  10-20 and 21-30 for Prompt 1 vs Prompt 7. Prompt 1 is a more explicit phrasing for the same prompt, whereas Prompt 7 is slightly vaguer. $\alpha = 0.2$}
    \label{fig:phrasing both bracket 1&7}
\end{figure}

\begin{figure}[h!]
    \centering
    \includegraphics[width=0.5\textwidth]{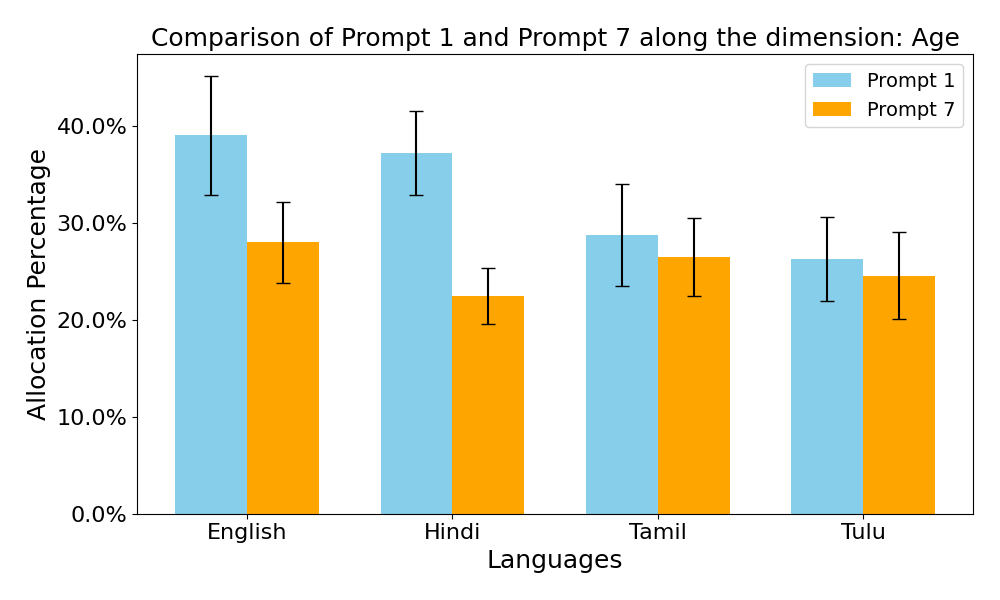}
    \caption{Plot showcasing the allocation percentages for the age values  10-20 for Prompt 1 vs Prompt 7. Prompt 1 is a more explicit phrasing for the same prompt, whereas Prompt 7 is slightly vaguer. $\alpha = 0.2$}
    \label{fig:phrasing first bracket 1&7}
\end{figure}

\begin{figure}[h!]
    \centering
    \includegraphics[width=0.5\textwidth]{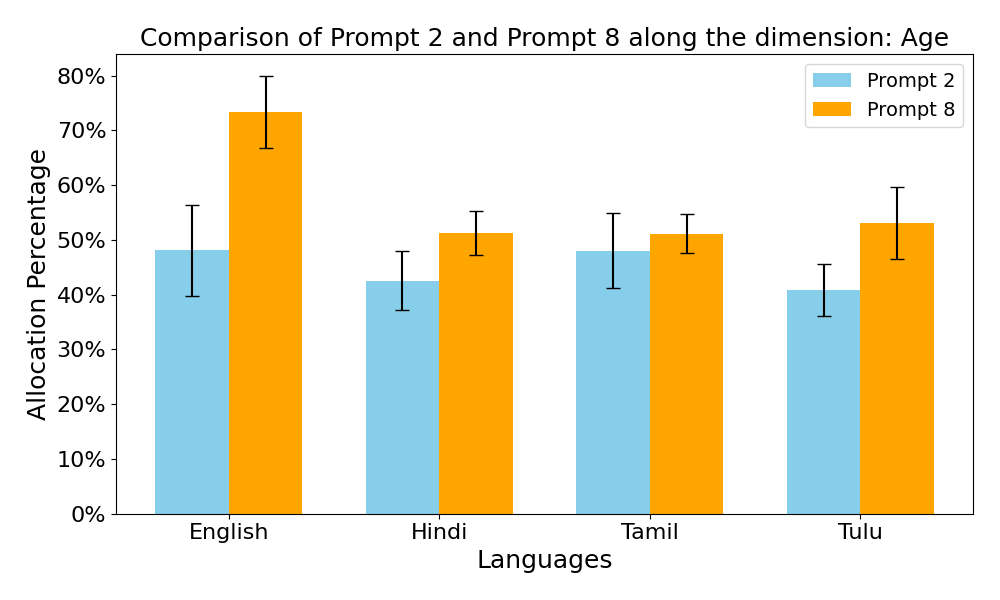}
    \caption{Plot showcasing the allocation percentages summed for the age values  31-40 and 41-50 for Prompt 2 vs Prompt 8. Prompt 8 is a more explicit phrasing for the same prompt, whereas Prompt 2 is slightly vaguer. $\alpha = 0.2$}
    \label{fig:phrasing both bracket 2&8}
\end{figure}

\begin{figure}[h!]
    \centering
    \includegraphics[width=0.5\textwidth]{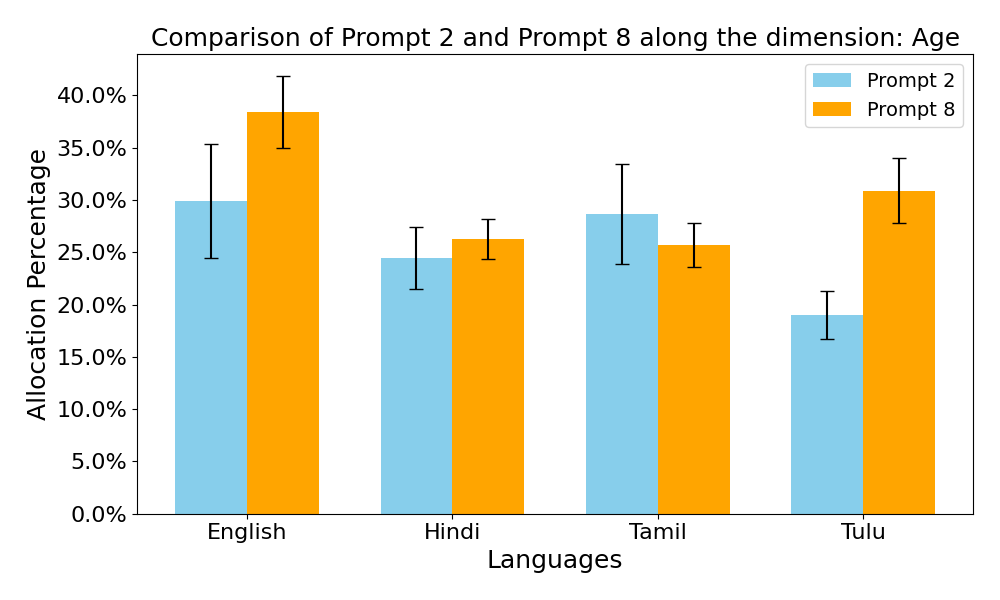}
    \caption{Plot showcasing the allocation percentages for the age values  31-40 for Prompt 2 vs Prompt 8. Prompt 8 is a more explicit phrasing for the same prompt, whereas Prompt 2 is slightly vaguer. $\alpha = 0.2$}
    \label{fig:phrasing first bracket 2&8}
\end{figure}

% Refer to plots [TBD]\\
We analyze the effects of phrasing on the performance of the algorithm by plotting the allocation percentage values across different languages for prompts that have been phrased differently. We observe from Figures \ref{fig:phrasing both bracket 1&7}, \ref{fig:phrasing first bracket 1&7}, \ref{fig:phrasing both bracket 2&8},  and \ref{fig:phrasing first bracket 2&8} that changing the phrasing has a noticeable impact on the allocations.  While the prompts are semantically the same, the allocations are noticeably different.

% Story [TBD]\\
% TODO: move all the five $\alpha=0.8$ plots to appendix.\\ %Done

% TBD - Also add ``Tulu being a low-resource language struggles often unless the goals are explicitly stated.''

\begin{figure}[h!]
    \centering
    \includegraphics[width=0.5\textwidth]{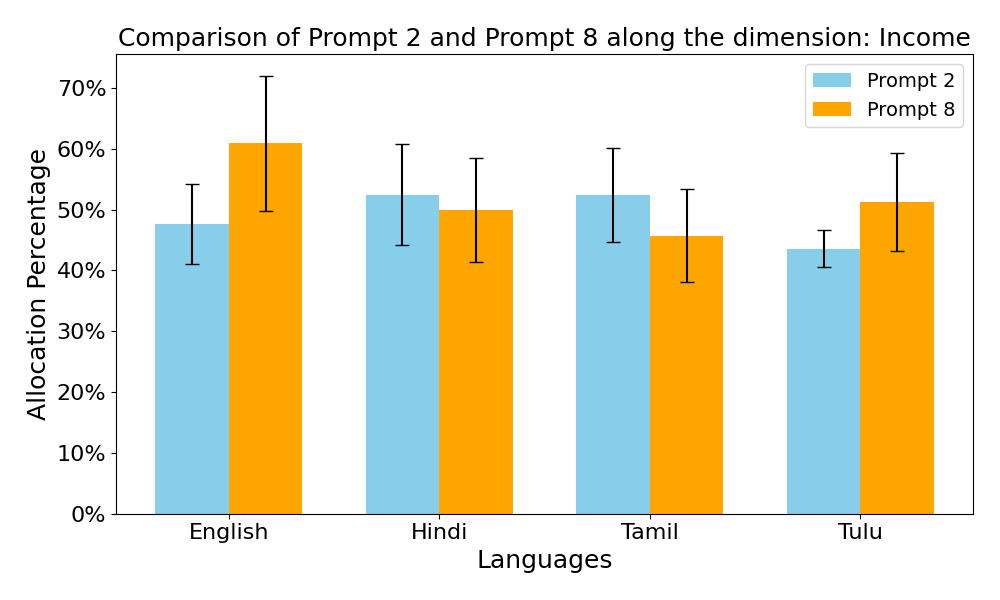}
    \caption{Plot showcasing the summed allocation percentages for the income values lower than 15000 for Prompt 2 vs Prompt 8. Prompt 8 is a more explicit phrasing for the same prompt, whereas Prompt 2 is slightly vaguer. $\alpha = 0.2$}
    \label{fig:income phrasing plot}
\end{figure}

\subsection{Result 3: Explicitly Stating your Goals Helps Improve Performance}
\label{res:result3}
We observe from Figures \ref{fig:phrasing both bracket 1&7}, \ref{fig:phrasing first bracket 1&7}, \ref{fig:phrasing both bracket 2&8},  and \ref{fig:phrasing first bracket 2&8} that explicitly stating our goals helps elicit better performance from the algorithm. In fact, in some instances, the improvement is quite significant. Additionally, from Figure \ref{fig:income phrasing plot}, we observe that even when the allocations are not as intended, explicitly stating the goal does improve allocations (in some cases, even pushing the needle to the point where the allocation could be considered desirable). It is something to be mindful of when using the DLM system. %as being explicit in stating the goals also require the translation to adhere to the norms and schema set in place in the LLM prompt.

In Appendix \ref{app:phrasing-plots-0.8}, we provide plots for $\alpha=0.8$ that further support our claims on the importance of phrasing and explicitly stating one's goals. 

\subsection{Result 4 : Performance Degrades with Increasing Prompt Complexity; But English is More Robust}
\label{res:result4}
\begin{figure}[h]
    \centering
    \includegraphics[width=0.5\textwidth]{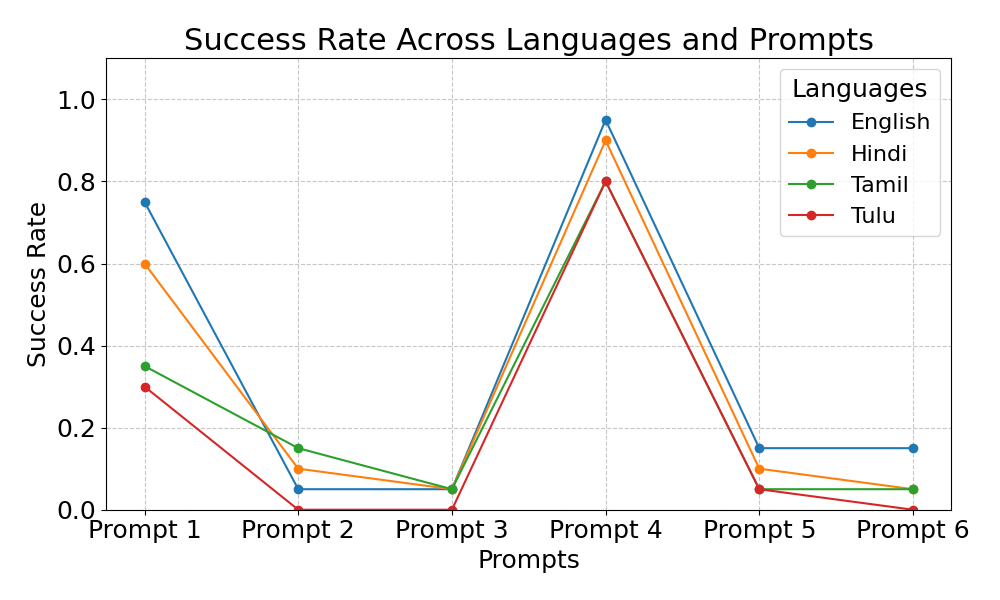}
    \caption{Plot measuring the success rate for different languages against increasing complexity of prompts for $\alpha = 0.2$.}
    \label{fig:complexity plot 0.2}
\end{figure}

\begin{figure}[h]
    \centering
    \includegraphics[width=0.5\textwidth]{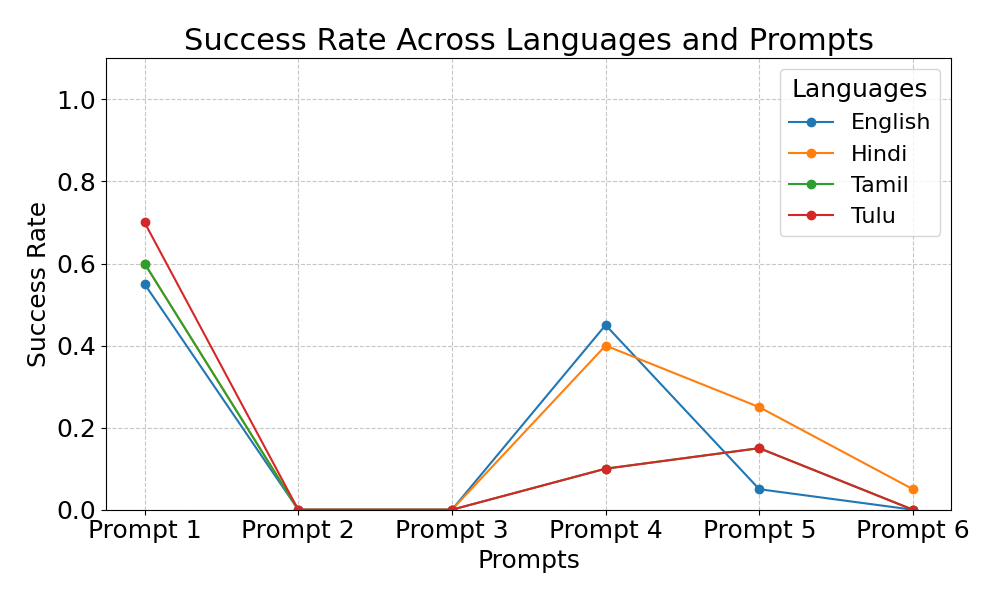}
    \caption{Plot measuring the success rate for different languages against increasing complexity of prompts for $\alpha = 0.8$}
    \label{fig:complexity plot 0.8}
\end{figure}

We analyzed the task success rate (the percentage of our runs that successfully carried out the task) against each prompt, with the complexity increasing as we moved along the x-axis. Here, we define success as an instance where the allocations have deviated from the relevant features to the point where the allocation is higher than what it would have been had the allocation been uniform across feature values. 
From the plots Figure \ref{fig:complexity plot 0.2} and Figure \ref{fig:complexity plot 0.8}, we see that as the complexity of the prompt increases, English, more often than not, performs better than the low resource languages. It is also to be noted that beyond a certain threshold of complexity (three feature allocation), performance collapses regardless of language.

In Appendix \ref{app:alloc-plots}, we have provided various allocation plots for the same, where we observe that English prompts frequently achieve the patterns of allocation intended by the goal prompt.

{\textbf{Note}}: {The very poor performance with  all languages for Prompts 2 and 3 is due to the failure to provide correct allocations for the features \textit{Income} and \textit{Times\_To\_Be\_Called}, both of them being ordered categorical features. We hypothesize that these types of features require explicit expression of the ranges, but further analysis is required for a definitive stance.}

From a deployability standpoint, the results of sections \ref{res:result2}, \ref{res:result3} and \ref{res:result4} pose interesting questions. The difference in performance for different languages might be a cause for concern, especially in use cases like resource allocation for public health. These disparities would lead to a difference in outcomes for users who use different languages, thereby leading to unfairness. If the algorithm is to be deployed for grassroots-level interventions, where one might see the language dimension come to the fore, this difference in performance for different languages may cause undesirable outcomes. While there are various possible directions one could take to improve the models so as to mitigate these issues, our results help show that these disparities exist and we must be aware of them and try to tackle them to create more equitable solutions. 

% Also bake in ``We observe that English consistently delivers on the intended consequences as compared to low resource languages''

% \subsection{Result 5: explictly worded prompts seem to peform better}
% Below we have plotted the allocations for various prompts for different languages \\

%  When the specific feature value is more explicitly mentioned in the prompt, the LLM seems to perform better than when it is indirect and the LLM has to infer it. 
    
%     For example, consider Figures [prompt-3-0.2] and [prompt-4-0.2]. In Figure [prompt-3-0.2], we see [TBD]. Whereas in Figure [prompt-4-0.2], where the prompt is much more explicit with respect to the feature values, we see [TBD].

%     As another example, in Figure [phrasing-plot], we see [TBD].

%     These lead us to believe that ... [TBD].

\section{Analysis 3: Fairness Across Multilingual Prompts }
\label{sec:analysis-3-fairness}
\begin{figure}[h!]
    \centering
    \includegraphics[width=0.5\textwidth]{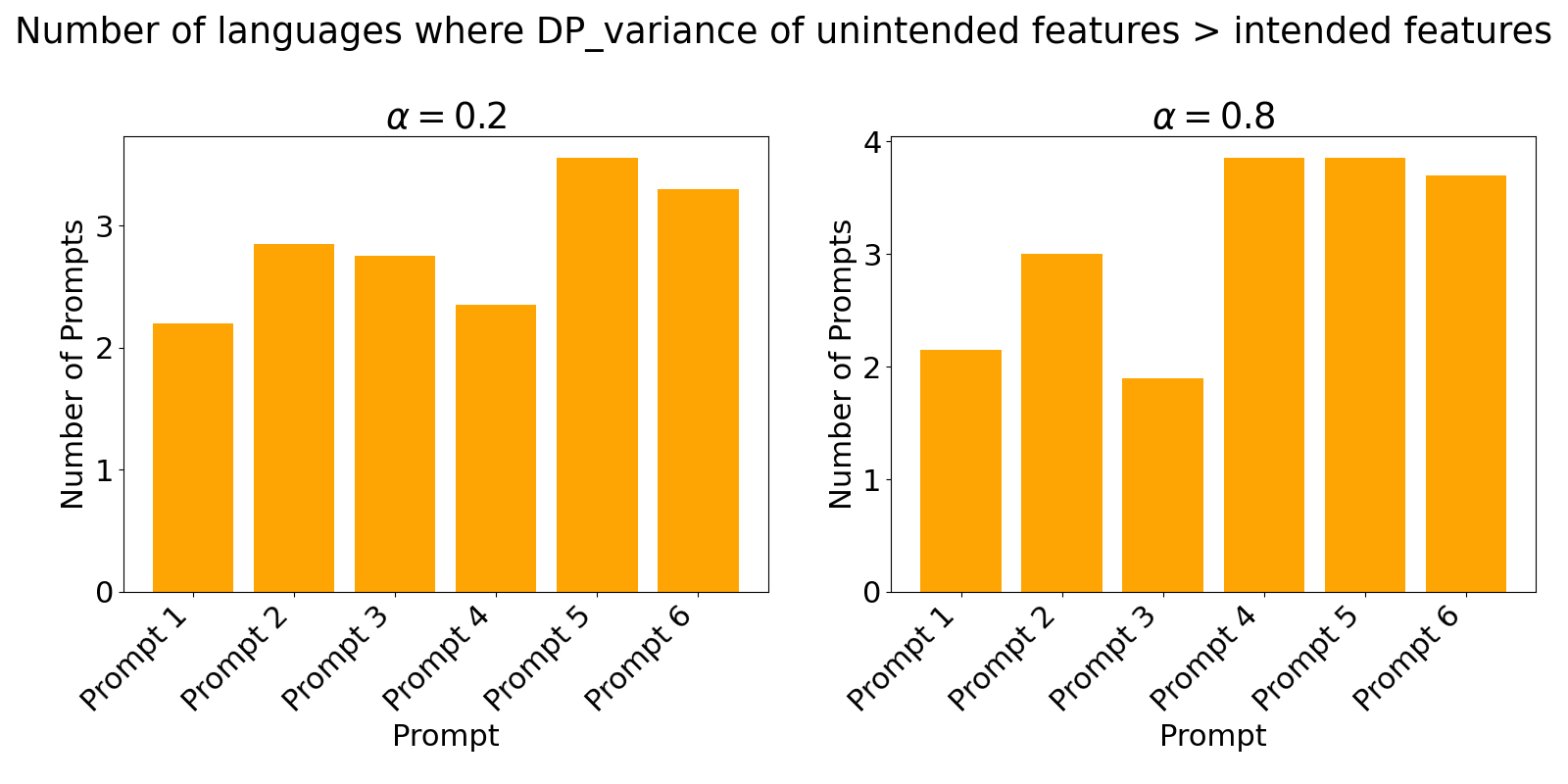}
    \caption{Average number of languages in each prompt where the \textit{DP\_variance} is higher in unintended features than in intended features (higher is more unfair)}
    \label{fig:prompt-comparison}
\end{figure}

Each prompt specifies how allocations are distributed across different features. For instance, \textit{Prompt 1} demands a higher allocation for individuals with lower age values. In this case, \textit{Age} is considered an `intended feature' for \textit{Prompt 1}, while all other features are `unintended'. Thus, a higher \textit{DP\_variance} in the unintended features indicates unfairness.
We plot two main counts as a measure of unfairness:
\begin{itemize}
    \item Absolute: We plot the number of prompts where \textit{DP\_variance} in unintended features is greater than a certain threshold for various values of thresholds.
    \item Relative: We plot the number of prompts where \textit{DP\_variance} in unintended features is greater than \textit{DP\_variance} in intended feature.
\end{itemize}
In Figure  \ref{fig:lang-comparison}, Figure \ref{fig:dp-variance-threshold-02}, Figure \ref{fig:dp-variance-threshold-08} and Figure \ref{fig:prompt-comparison}, we plot the average counts across 20 runs. 

\subsection{Result 5: Low Resource Languages are More Likely to Introduce Unfairness}

\begin{figure}[h!]
    \centering
    \includegraphics[width=0.5\textwidth]{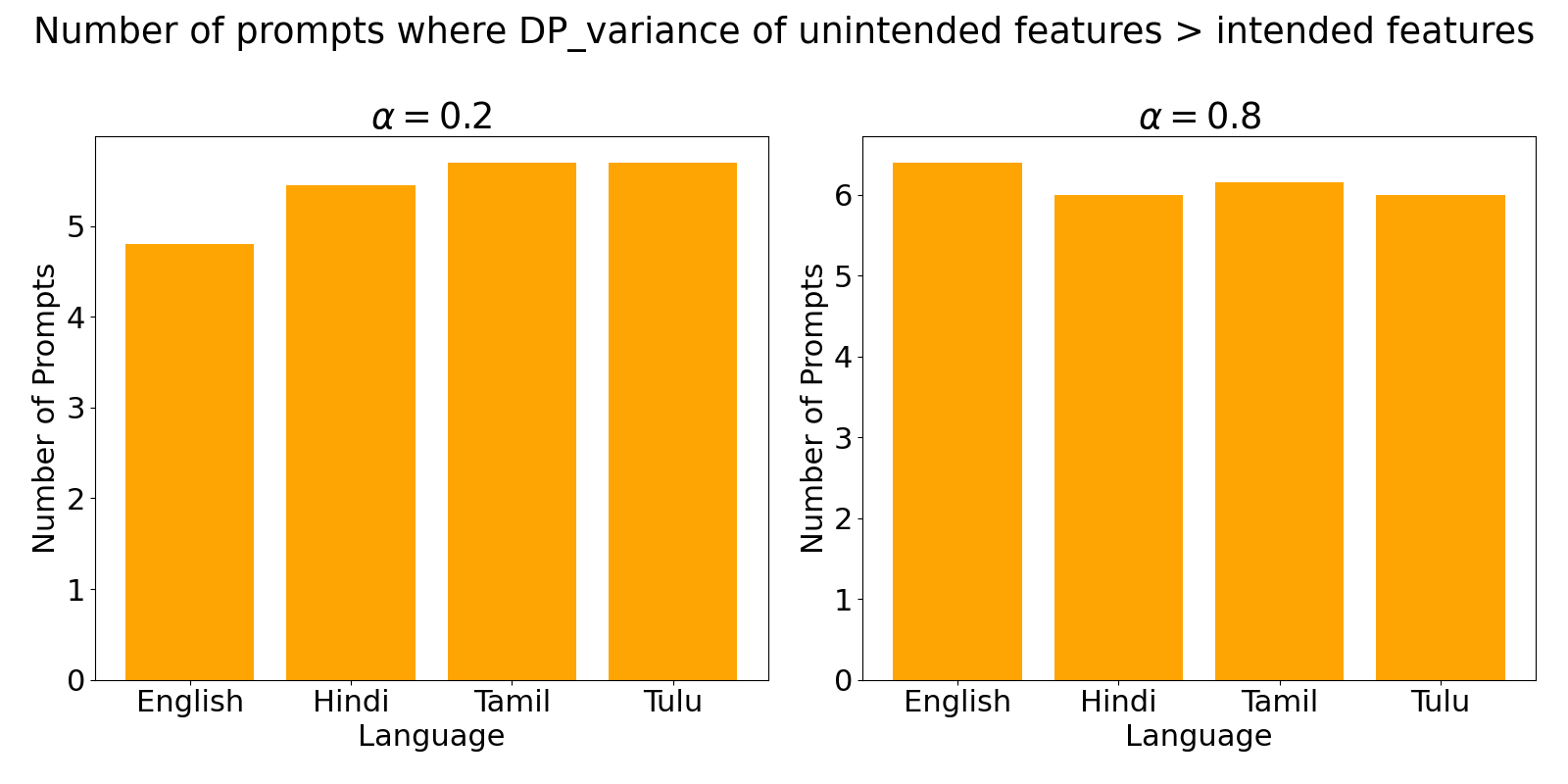}
    \caption{Number of prompts in each language where the \textit{DP\_variance} is higher in unintended features than in intended features (higher is more unfair)}
    \label{fig:lang-comparison}
\end{figure}

\begin{figure}[h!]
    \centering
    \includegraphics[width=0.5\textwidth]{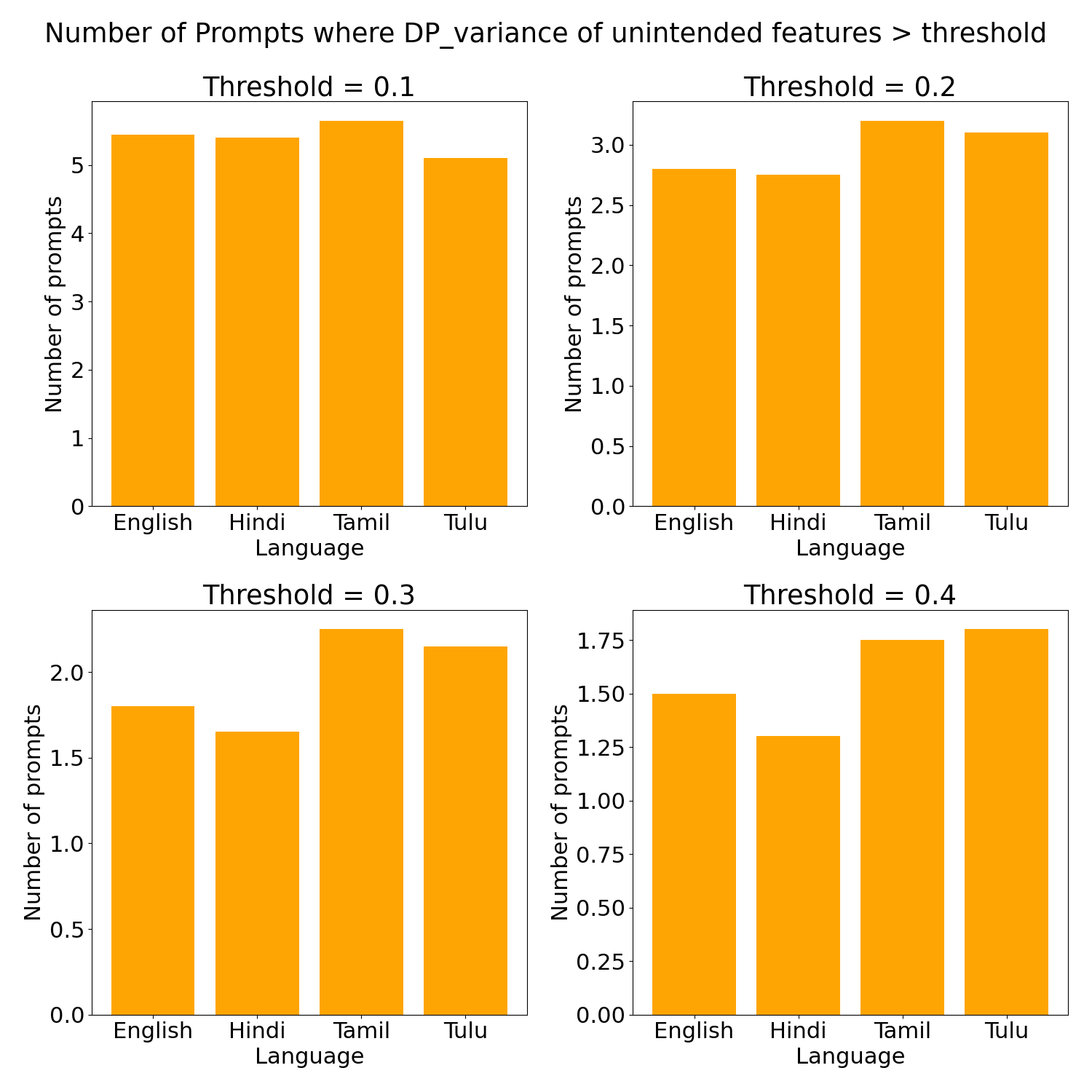}
    \caption{Average number of prompts where the \textit{DP\_variance} of unintended features is greater than a threshold value (higher is more unfair) for $\alpha = 0.2$}
    \label{fig:dp-variance-threshold-02}
\end{figure}

\begin{figure}[h!]
    \centering
    \includegraphics[width=0.5\textwidth]{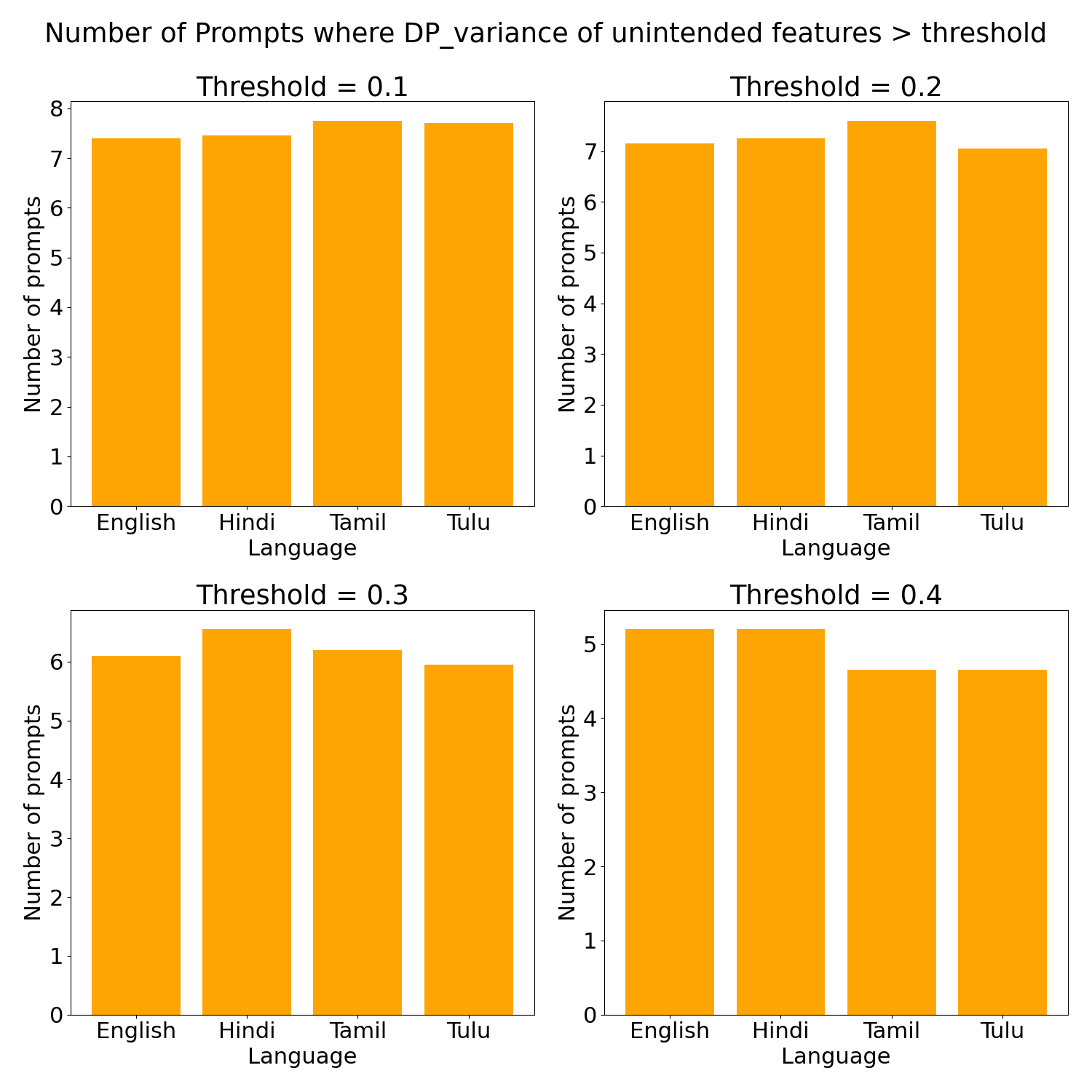}
    \caption{Average number of prompts where the \textit{DP\_variance} of unintended features is greater than a threshold value (higher is more unfair) for $\alpha = 0.8$}

    \label{fig:dp-variance-threshold-08}
\end{figure}

From the plots in Figure  \ref{fig:lang-comparison}, Figure \ref{fig:dp-variance-threshold-02} and Figure \ref{fig:dp-variance-threshold-08}, we see that for $\alpha = 0.2$, low-resource languages like Tulu are more unfair as evident from the high count values. This is not as pronounced in $\alpha = 0.8$ since the high \textit{DP\_variance} in the intended feature propagates to the unintended features due to high conditional correlation.

\subsection{Result 6: More Complex Prompts Tend to Result in More Unfairness}
Prompts become more complex when the number of intended features increases, or when the prompt is not direct and requires a step of reasoning to fully understand it.
Prompts 5 and 6 are more complex than prompts 1, 2 and 3 since they have more number of intended features and are harder to interpret. From Figure \ref{fig:prompt-comparison}, prompts 5 and 6 seem to be more unfair than prompts 1, 2 and 3, supporting our claim that more complex prompts are more likely to be unfair.

% \subsection{Result 7: Connection of fairness with task performance}
% [Plots - TBD]

% Refe our paper, including, for example, this instructions file, unused graphics files, style files, additional material sent for the purpose of the paper review, intermediate build files and so forth.

\section{Directions for Future Work}
\label{sec:future-work}
We propose a few directions for future work that can potentially mitigate some of the issues that came up in our experiments and analyses. 

\paragraph{Prompt rewriting component}
We could have a separate `prompt rewriting' component in the DLM algorithm that takes the user prompt and rewrites it before sending it to the rest of the DLM algorithm. Specifically, this component could rewrite prompts with the right phrasing and possibly also make feature values desired by the user more explicit in the prompt. The results in our experimental sections indicate that both of these could potentially improve task performance.

\paragraph{Improved reflection component}
Ideally, even if one of the multiple proposed reward functions is good, the LLM tasked with reflection is supposed to choose it. However, this frequently does not happen, as we can see in Section \ref{sec:analysis-1-acceptable-reward}. Anecdotally as well, we saw that the LLM that does the evaluation sometimes picks a worse reward function even if a much better proposal is available. One possible direction to investigate is to see if we can improve the reflection LLM to increase the chance that it picks the best reward function from multiple proposals. 

\paragraph{Prompt guardrails} 
Given that the system is intended to be used widely in critical settings such as public health, it is important to ensure that malicious users do not introduce undesirable biases through the prompt. We suggest that the system, before deployment, has the right guardrails in place to prevent such issues. One direction to explore is to have some kind of constitution \cite{bai2022constitutionalaiharmlessnessai} against which the prompts can be evaluated. 

\section{Conclusion}
In this paper, we study the effects on both task performance and fairness when the DLM algorithm \cite{behari2024decisionlanguagemodeldlmdynamic} is prompted in non-English languages. Our results show that the reward function proposals are better when prompted in English. Interestingly, we see that the way a prompt is worded (even if there is no change in meaning) affects task performance. Furthermore, performance degrades with prompt complexity, but it is more robust with English prompts than with lower-resource languages. Finally, our results show that prompts in lower-resource languages and more complex prompts are both more likely to create unfairness along unintended dimensions.

\appendix
\section{Appendix: Detailed Allocation Plots}
\label{app:alloc-plots}
\begin{figure*}[t]%[h!]
    \centering
    \includegraphics[width=\textwidth]{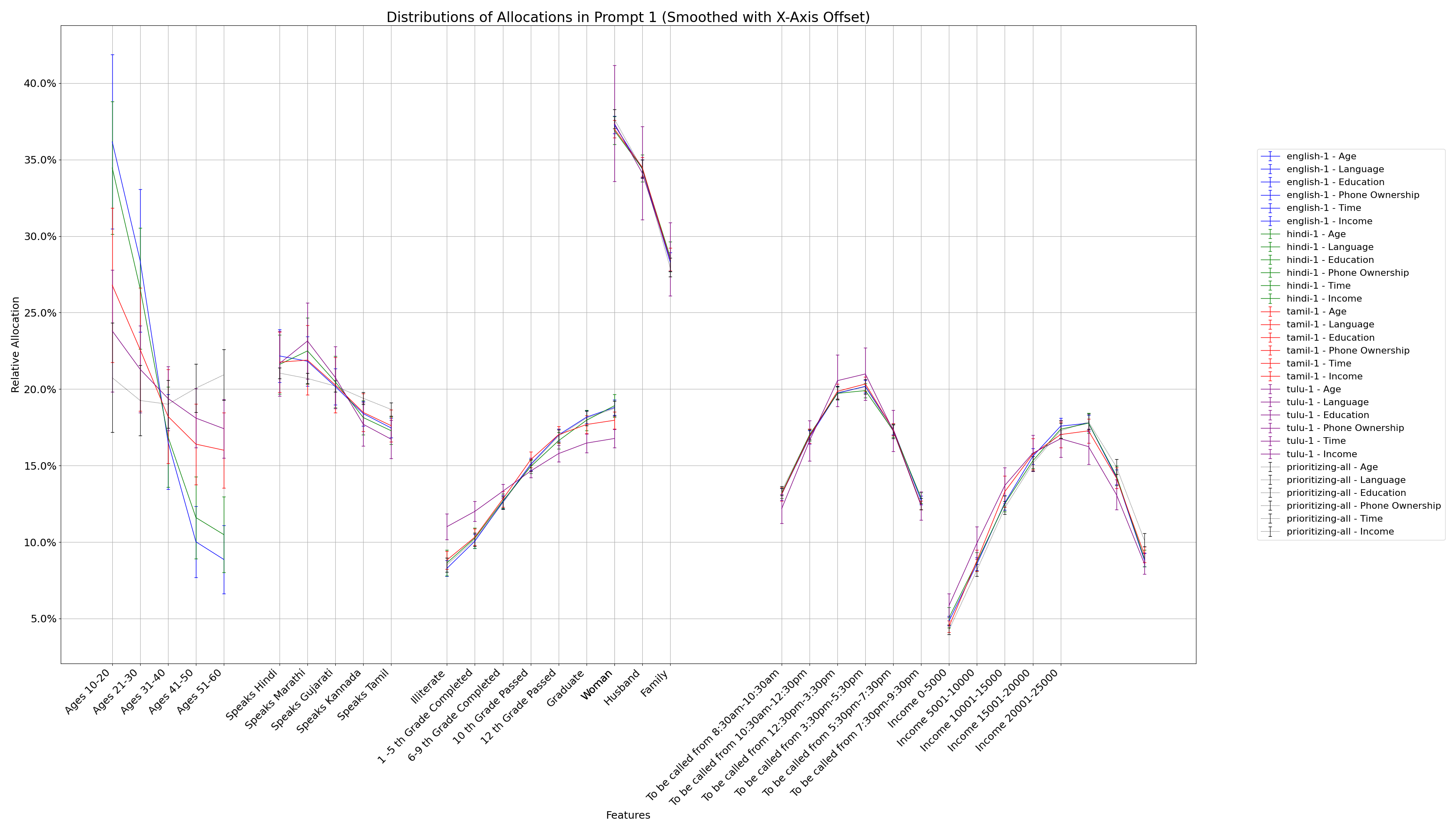}
    \caption{Allocation for Prompt 1, $\alpha = 0.2$}
    \label{fig:app-alloc-first}
\end{figure*}

\begin{figure*}[t]%[h!]
    \centering
    \includegraphics[width=\textwidth]{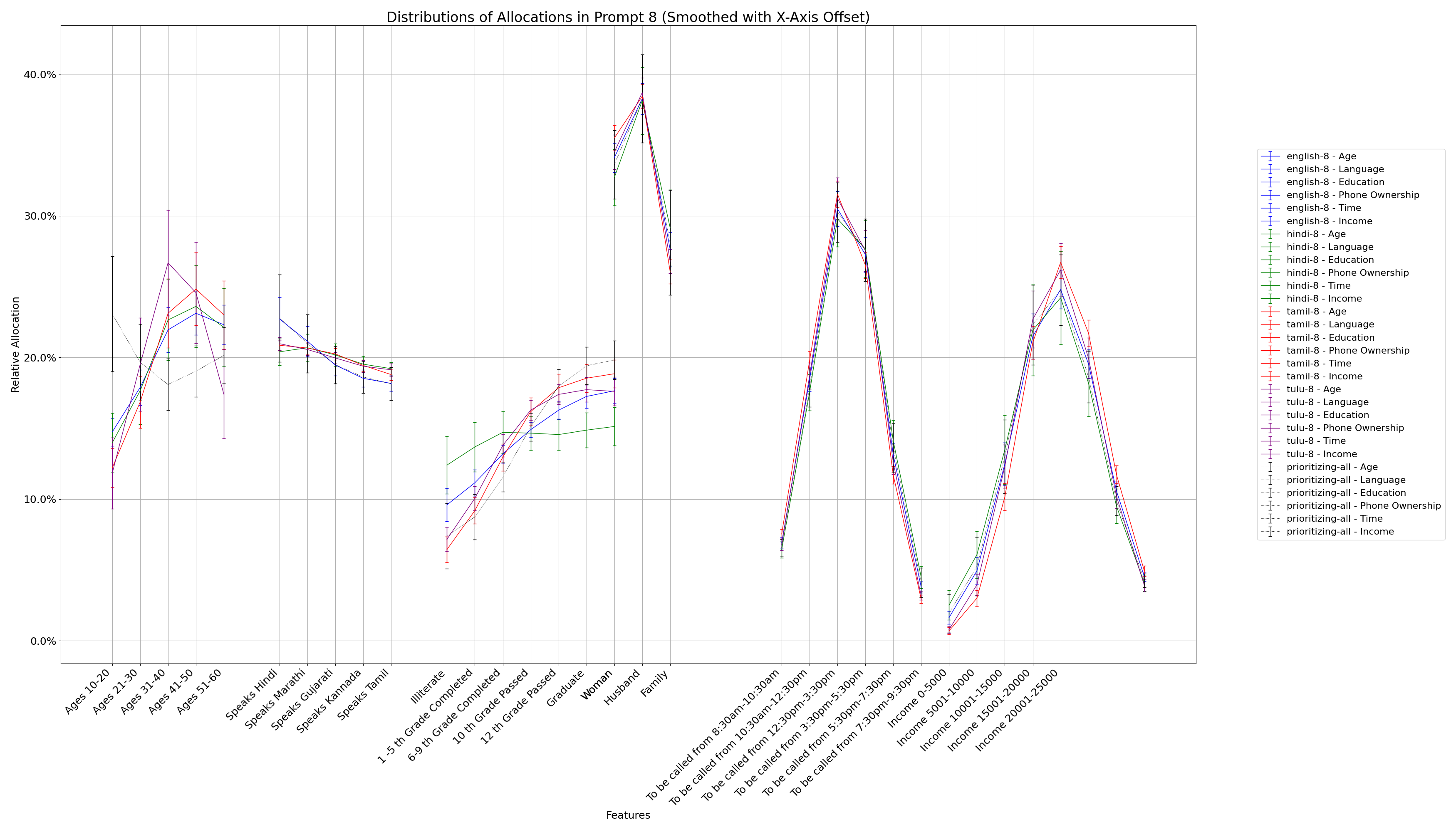}
    \caption{Allocation for Prompt 8, $\alpha = 0.8$}
    \label{fig:app-alloc-last}
\end{figure*}

Figures \ref{fig:app-alloc-first} through \ref{fig:app-alloc-last} for full allocation plots.

\begin{figure*}[t]%[h!]
    \centering
    \includegraphics[width=\textwidth]{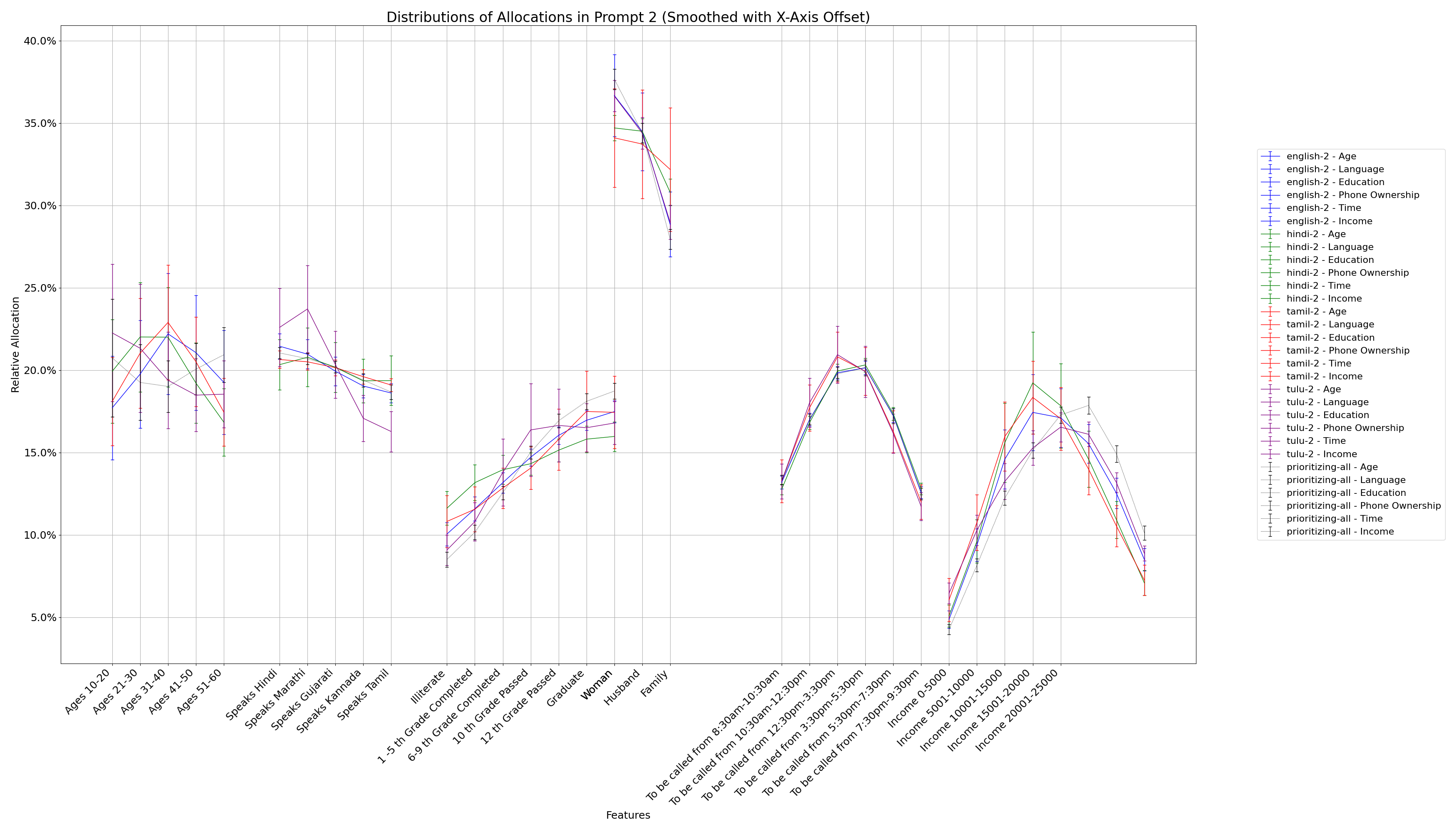}
    \caption{Allocation for Prompt 2, $\alpha = 0.2$}
    \label{fig:example}
\end{figure*}

\begin{figure*}[t]%[h!]
    \centering
    \includegraphics[width=\textwidth]{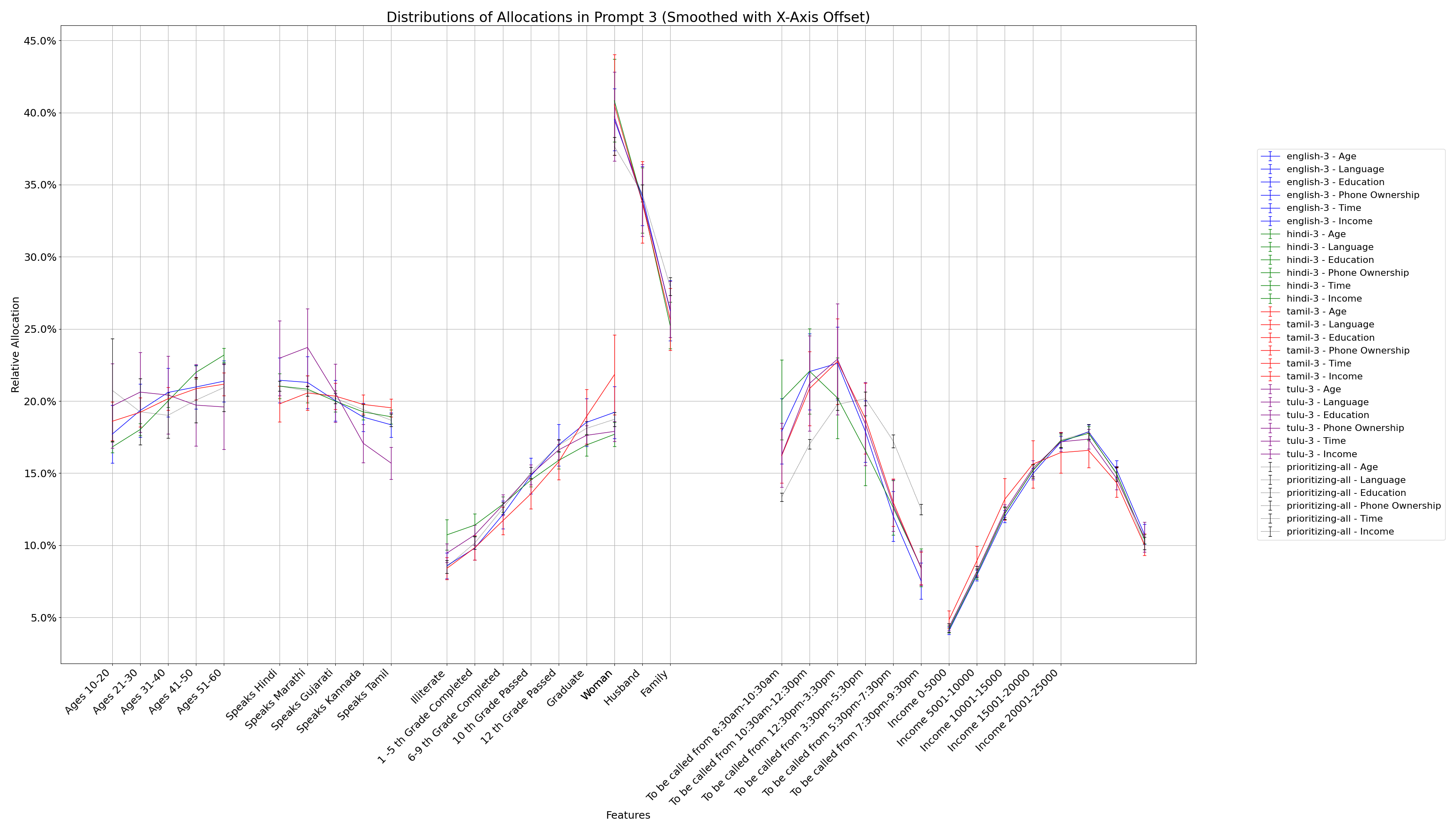}
    \caption{Allocation for Prompt 3, $\alpha = 0.2$}
    \label{fig:example}
\end{figure*}

\begin{figure*}[t]%[h!]
    \centering
    \includegraphics[width=\textwidth]{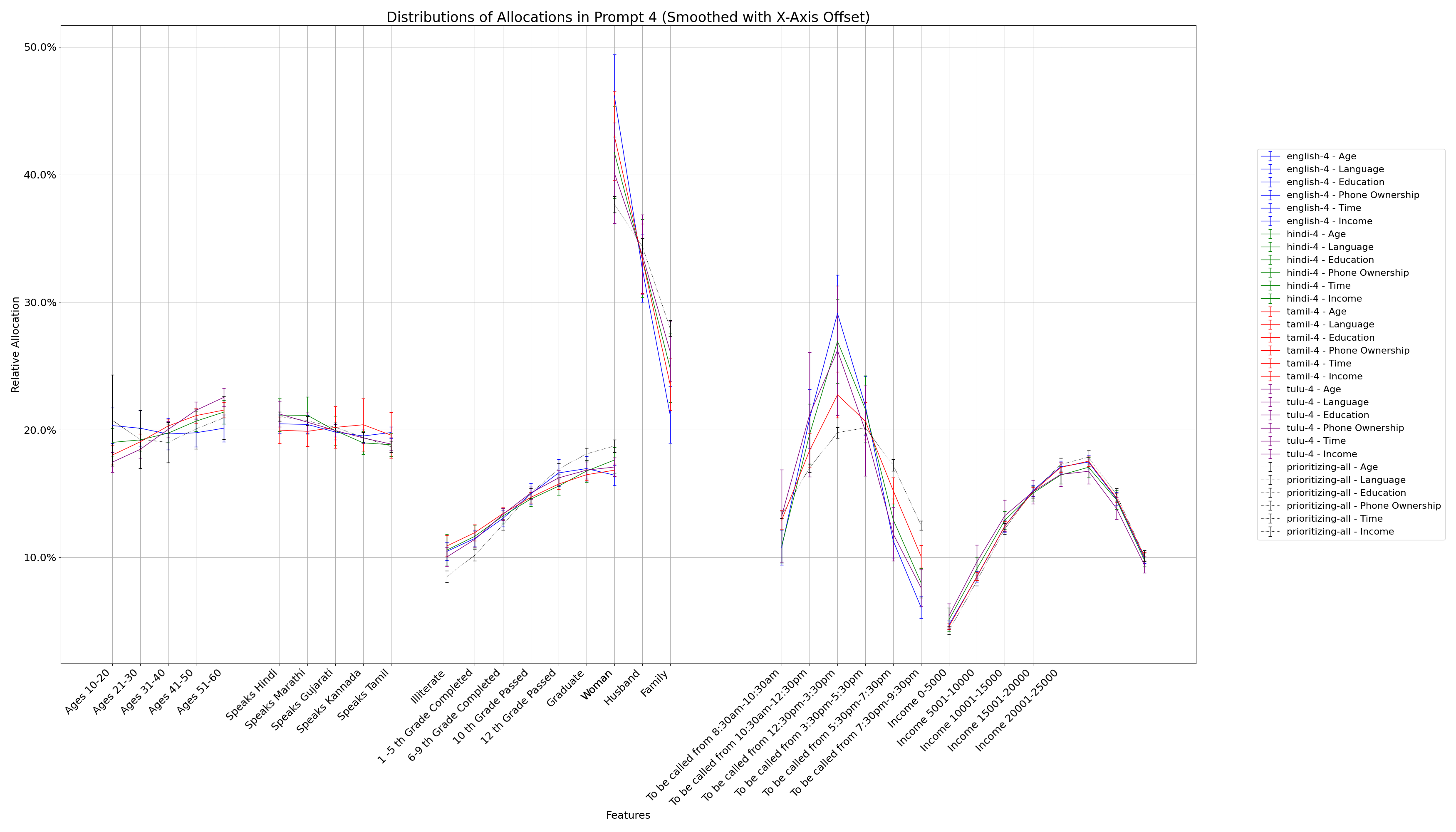}
    \caption{Allocation for Prompt 4, $\alpha = 0.2$}
    \label{fig:example}
\end{figure*}

\begin{figure*}[t]%[h!]
    \centering
    \includegraphics[width=\textwidth]{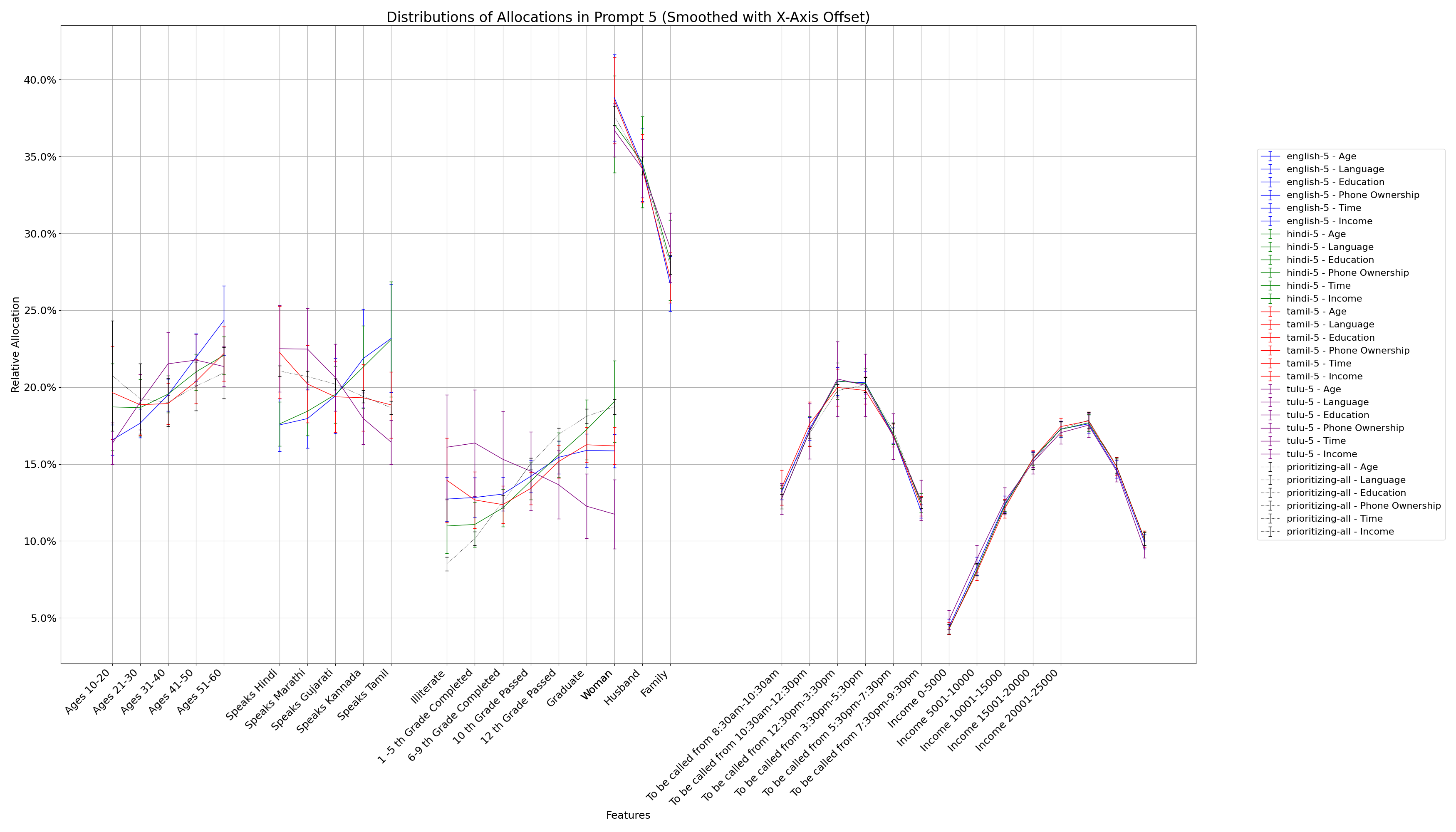}
    \caption{Allocation for Prompt 5, $\alpha = 0.2$}
    \label{fig:example}
\end{figure*}

\begin{figure*}[t]%[h!]
    \centering
    \includegraphics[width=\textwidth]{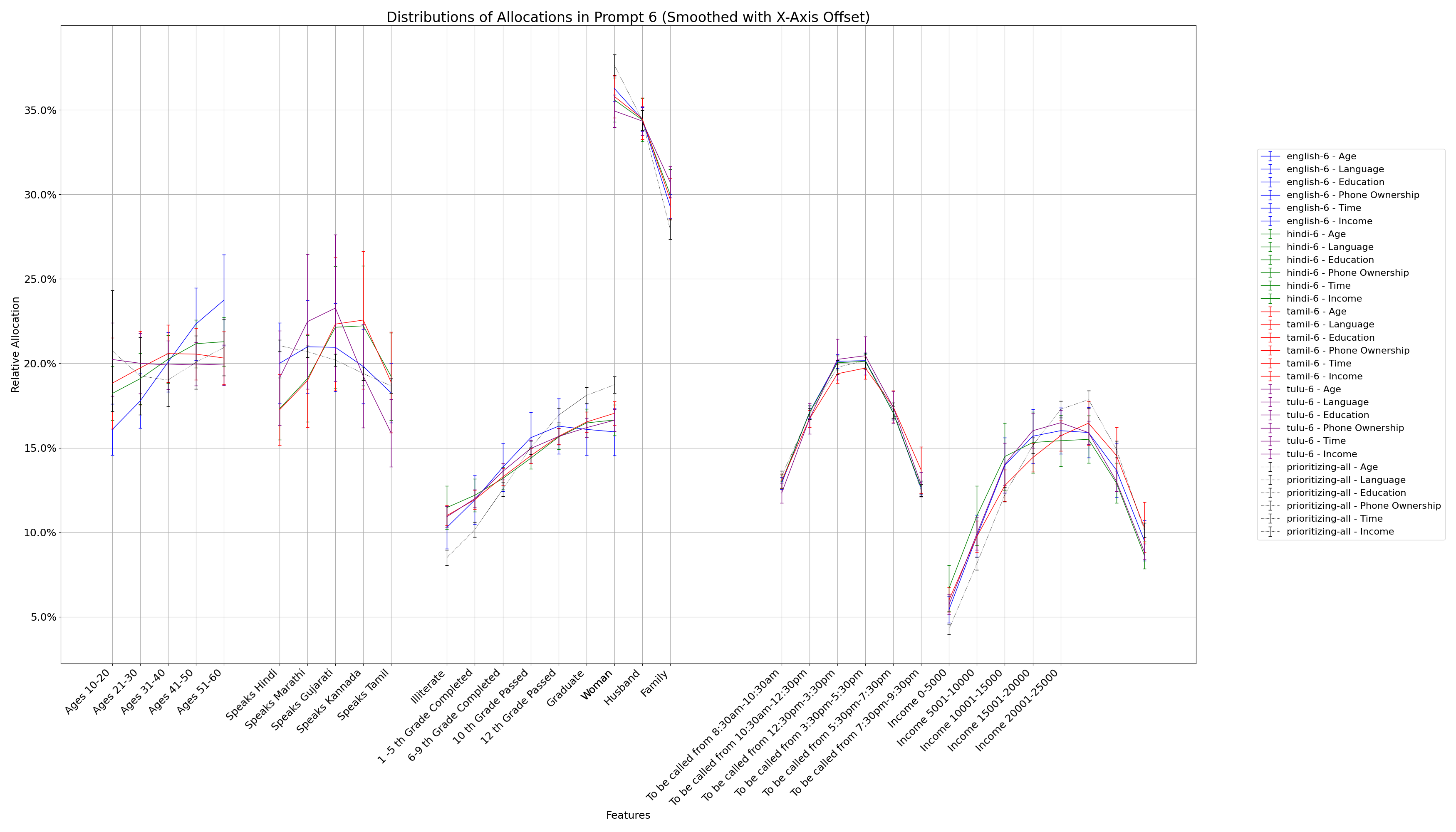}
    \caption{Allocation for Prompt 6, $\alpha = 0.2$}
    \label{fig:example}
\end{figure*}

\begin{figure*}[t]%[h!]
    \centering
    \includegraphics[width=\textwidth]{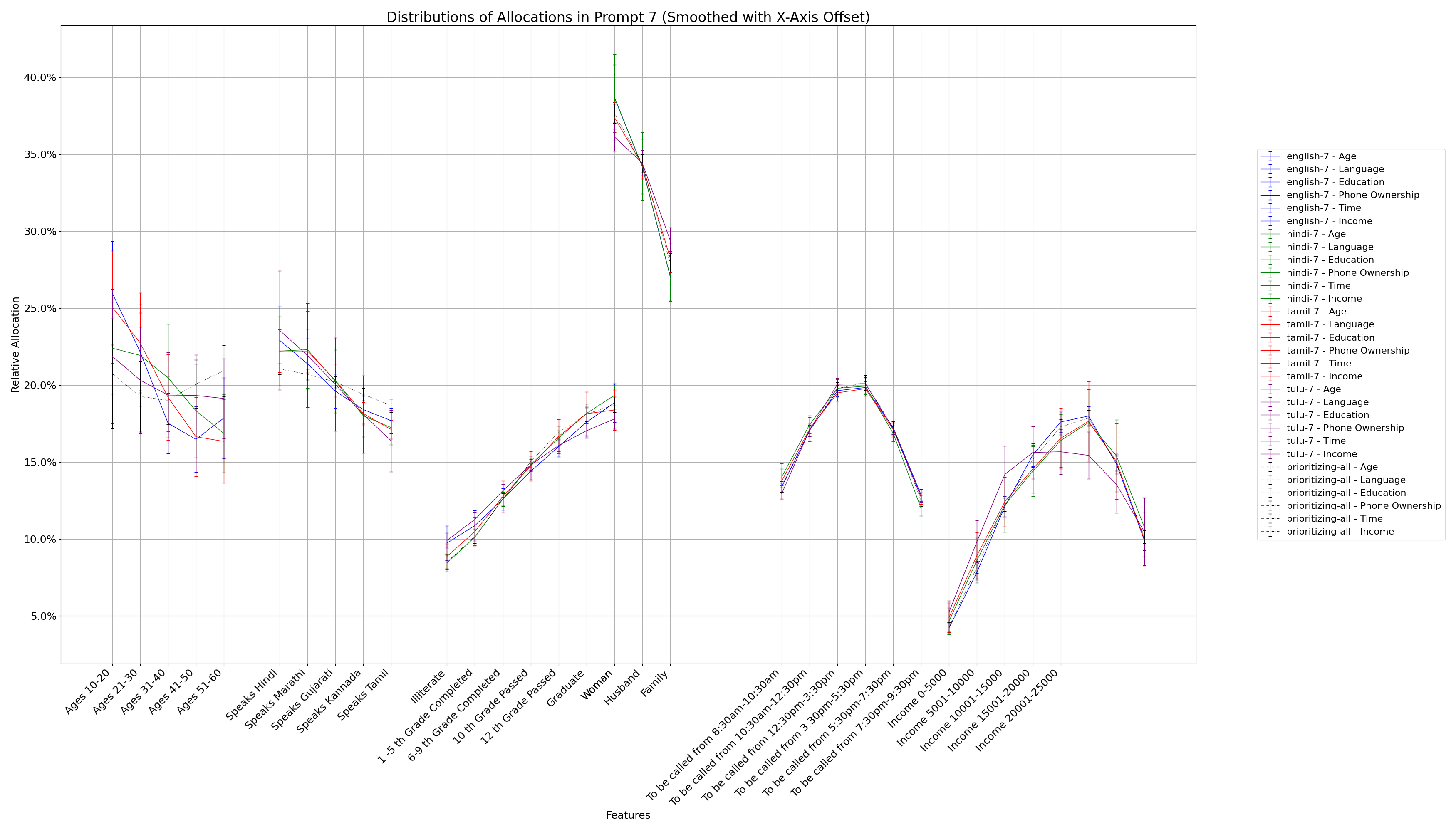}
    \caption{Allocation for Prompt 7, $\alpha = 0.2$}
    \label{fig:example}
\end{figure*}

\begin{figure*}[t]%[h!]
    \centering
    \includegraphics[width=\textwidth]{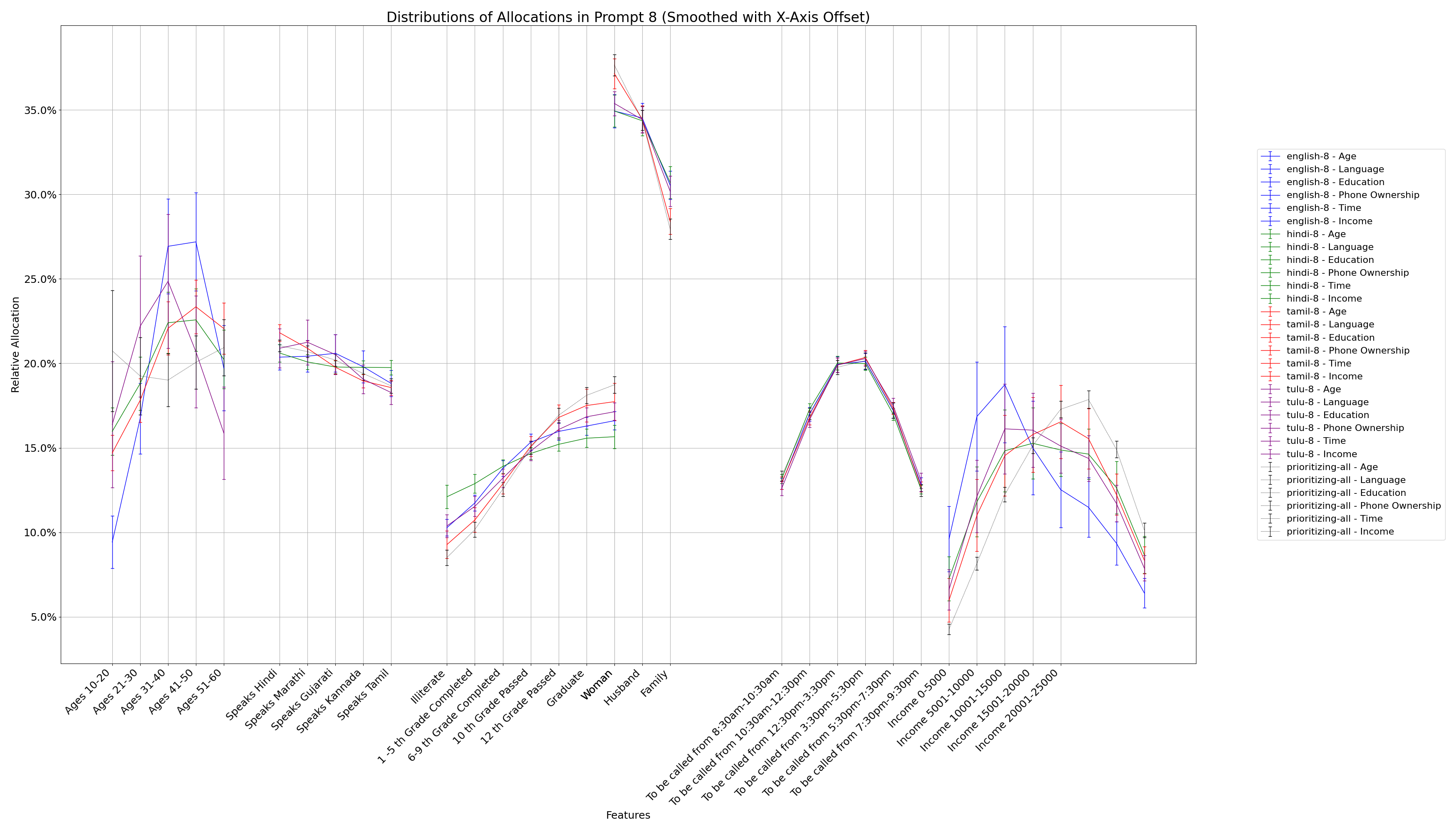}
    \caption{Allocation for Prompt 8, $\alpha = 0.2$}
    \label{fig:example}
\end{figure*}

\begin{figure*}[t]%[h!]
    \centering
    \includegraphics[width=\textwidth]{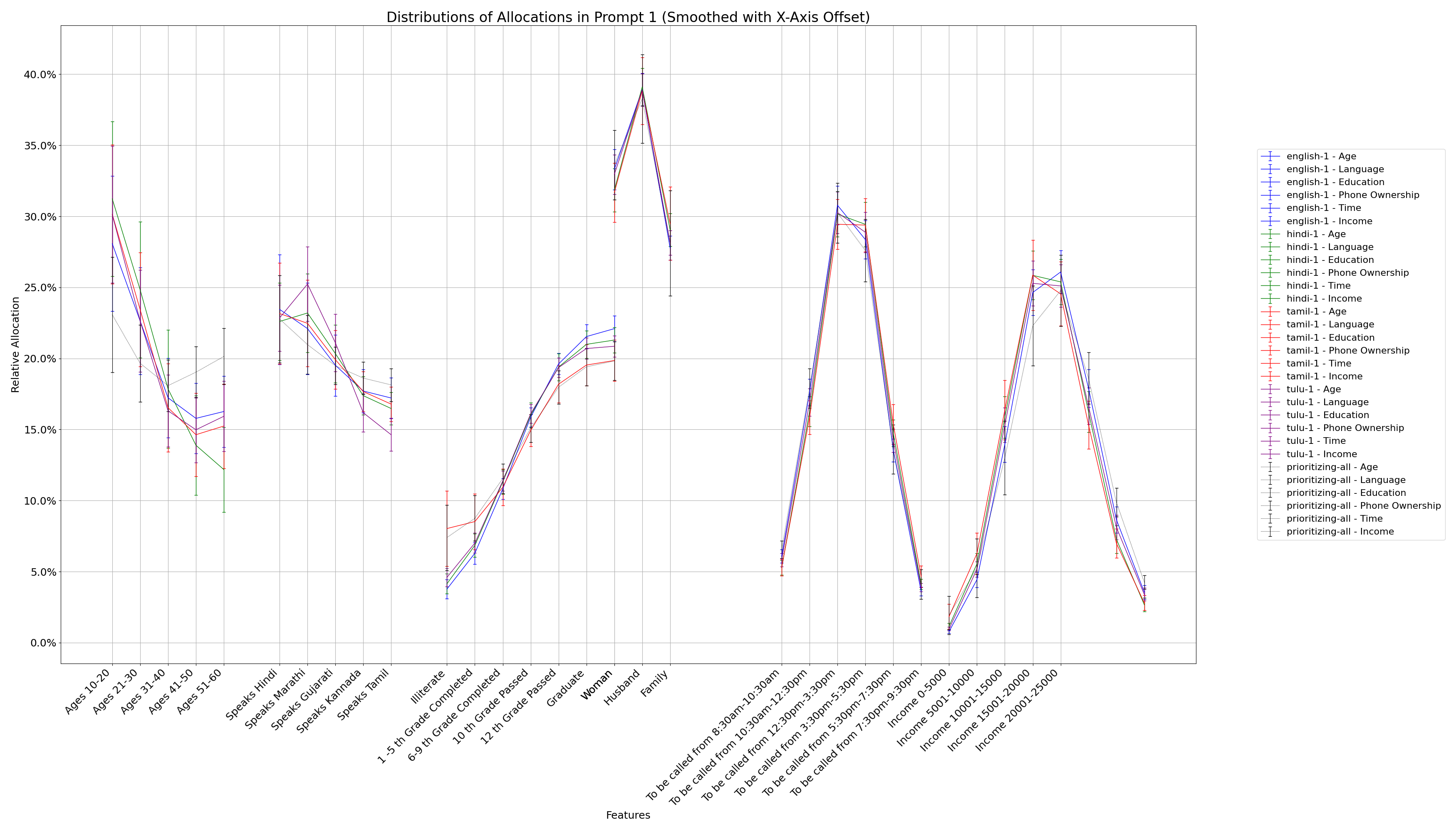}
    \caption{Allocation for Prompt 1, $\alpha = 0.8$}
    \label{fig:example}
\end{figure*}

\begin{figure*}[t]%[h!]
    \centering
    \includegraphics[width=\textwidth]{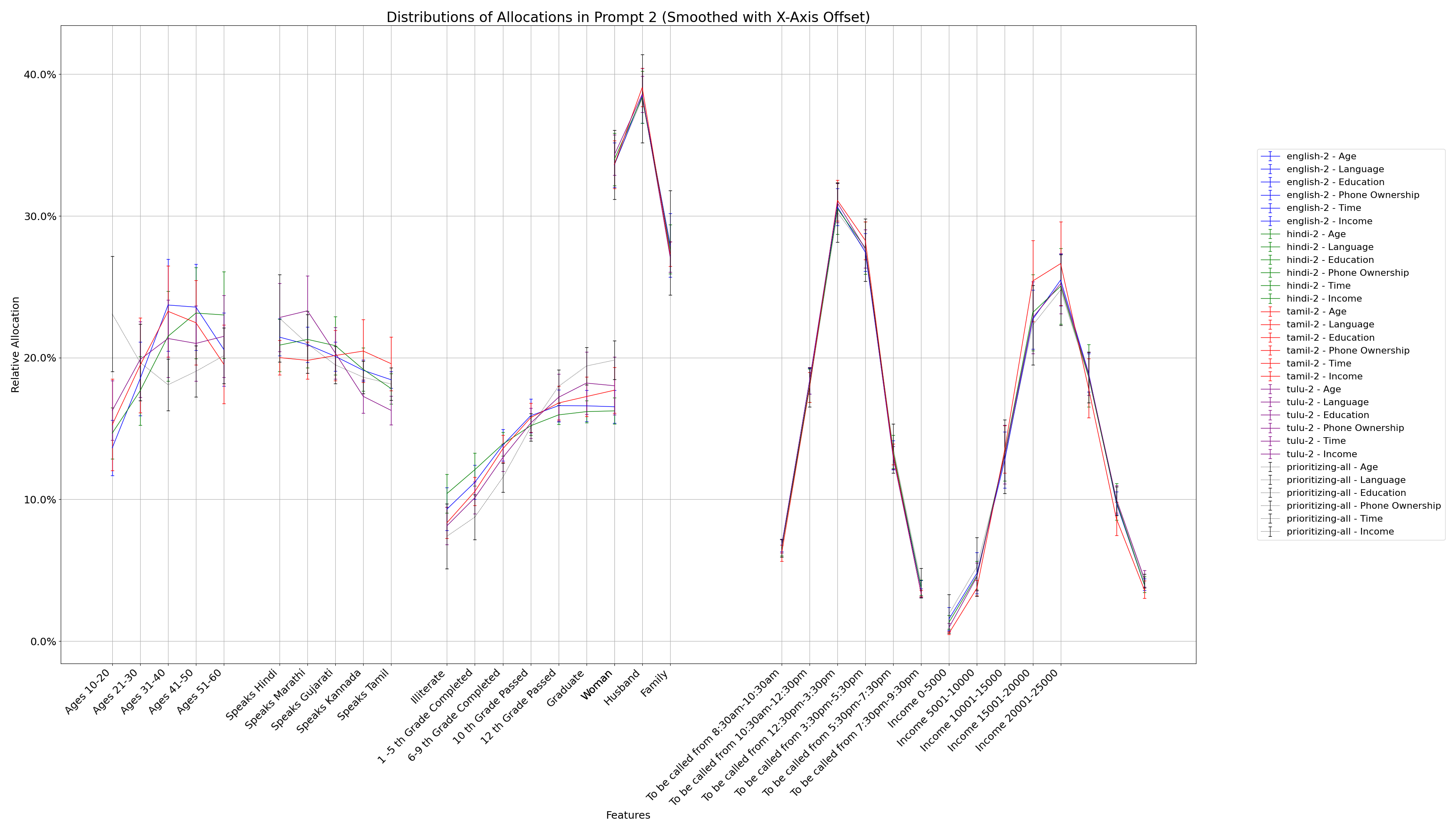}
    \caption{Allocation for Prompt 2, $\alpha = 0.8$}
    \label{fig:example}
\end{figure*}

\begin{figure*}[t]%[h!]
    \centering
    \includegraphics[width=\textwidth]{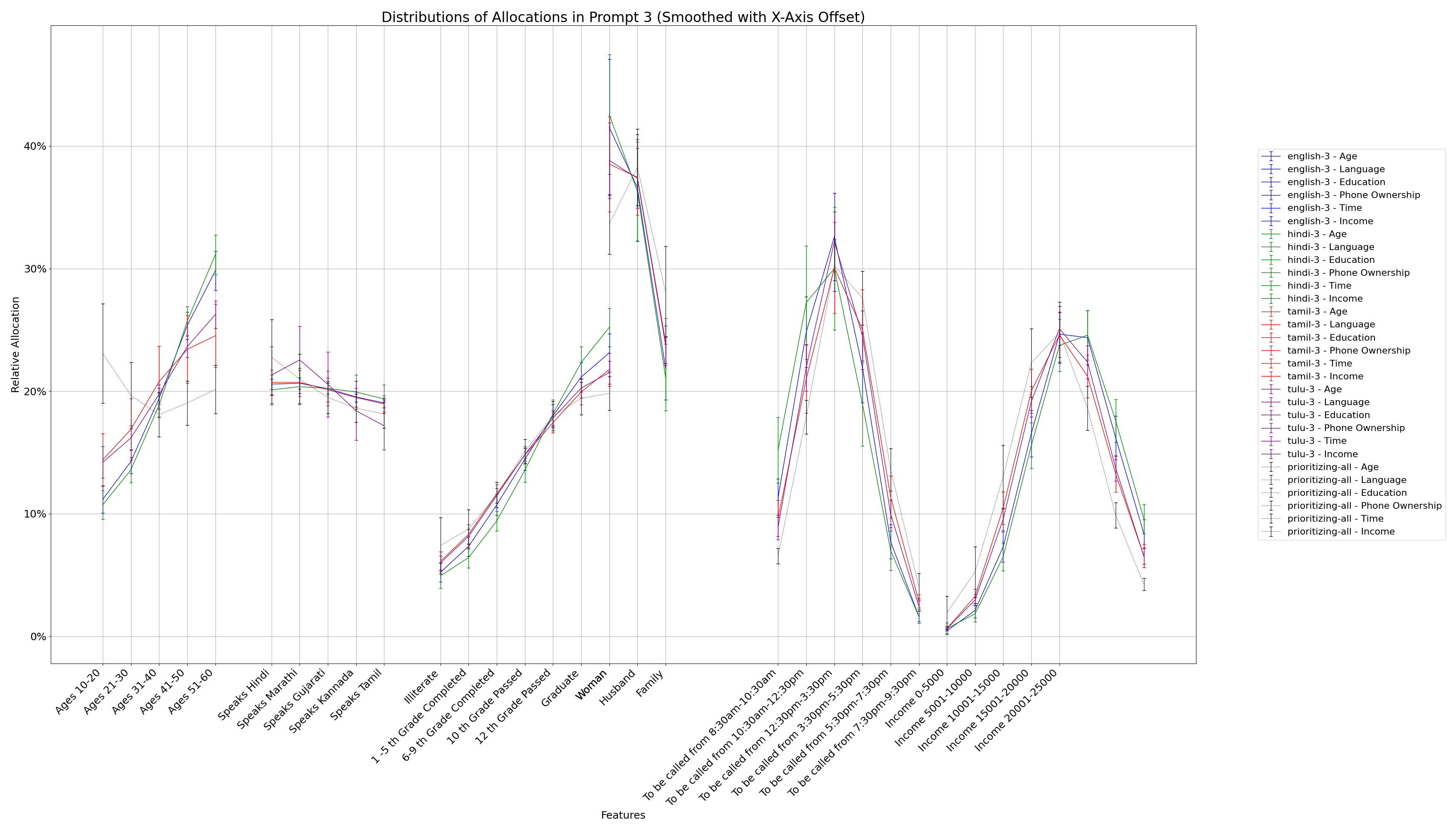}
    \caption{Allocation for Prompt 3, $\alpha = 0.8$}
    \label{fig:example}
\end{figure*}

\begin{figure*}[t]%[h!]
    \centering
    \includegraphics[width=\textwidth]{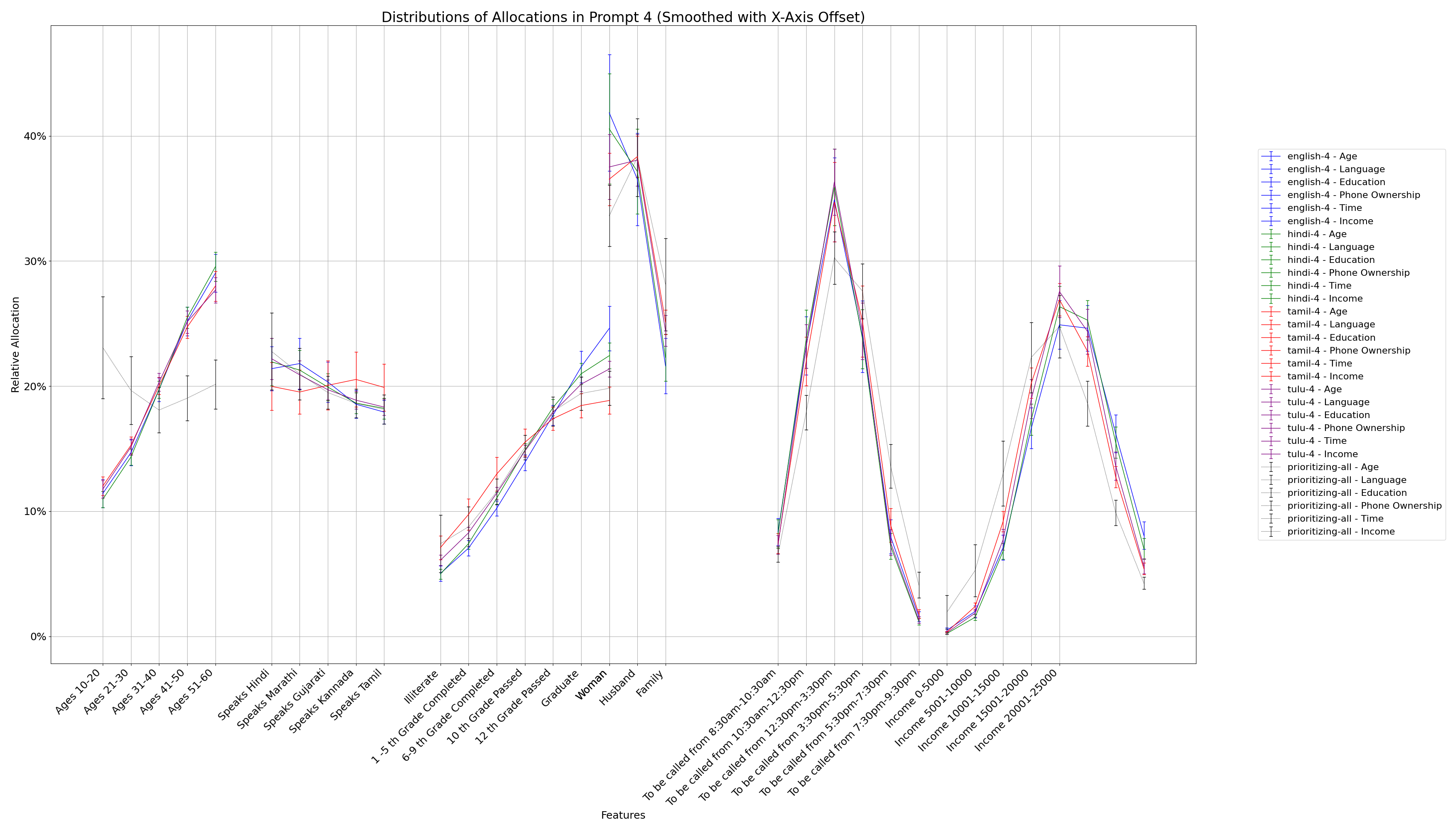}
    \caption{Allocation for Prompt 4, $\alpha = 0.8$}
    \label{fig:example}
\end{figure*}

\begin{figure*}[t]%[h!]
    \centering
    \includegraphics[width=\textwidth]{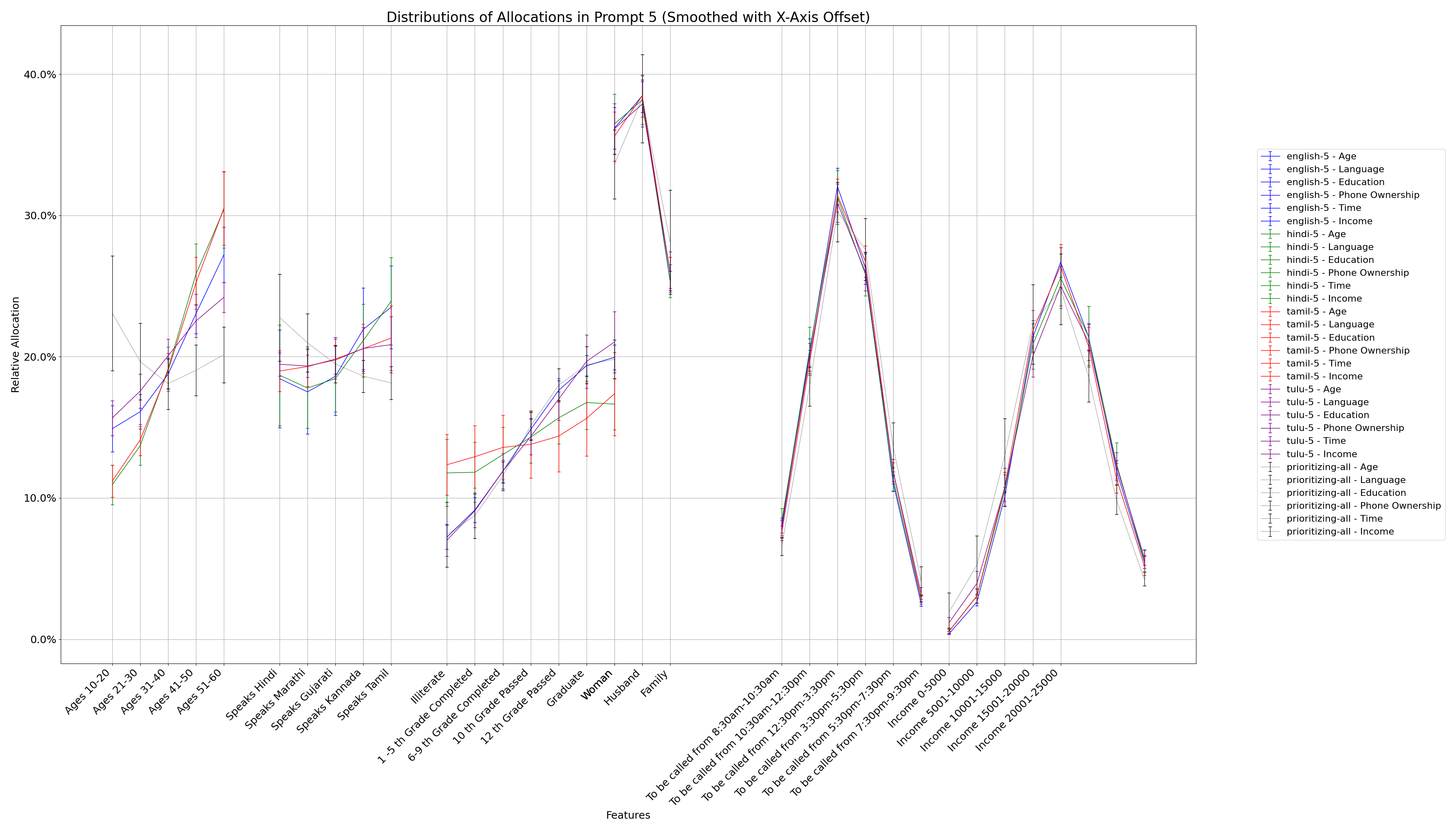}
    \caption{Allocation for Prompt 5, $\alpha = 0.8$}
    \label{fig:example}
\end{figure*}

\begin{figure*}[t]%[h!]
    \centering
    \includegraphics[width=\textwidth]{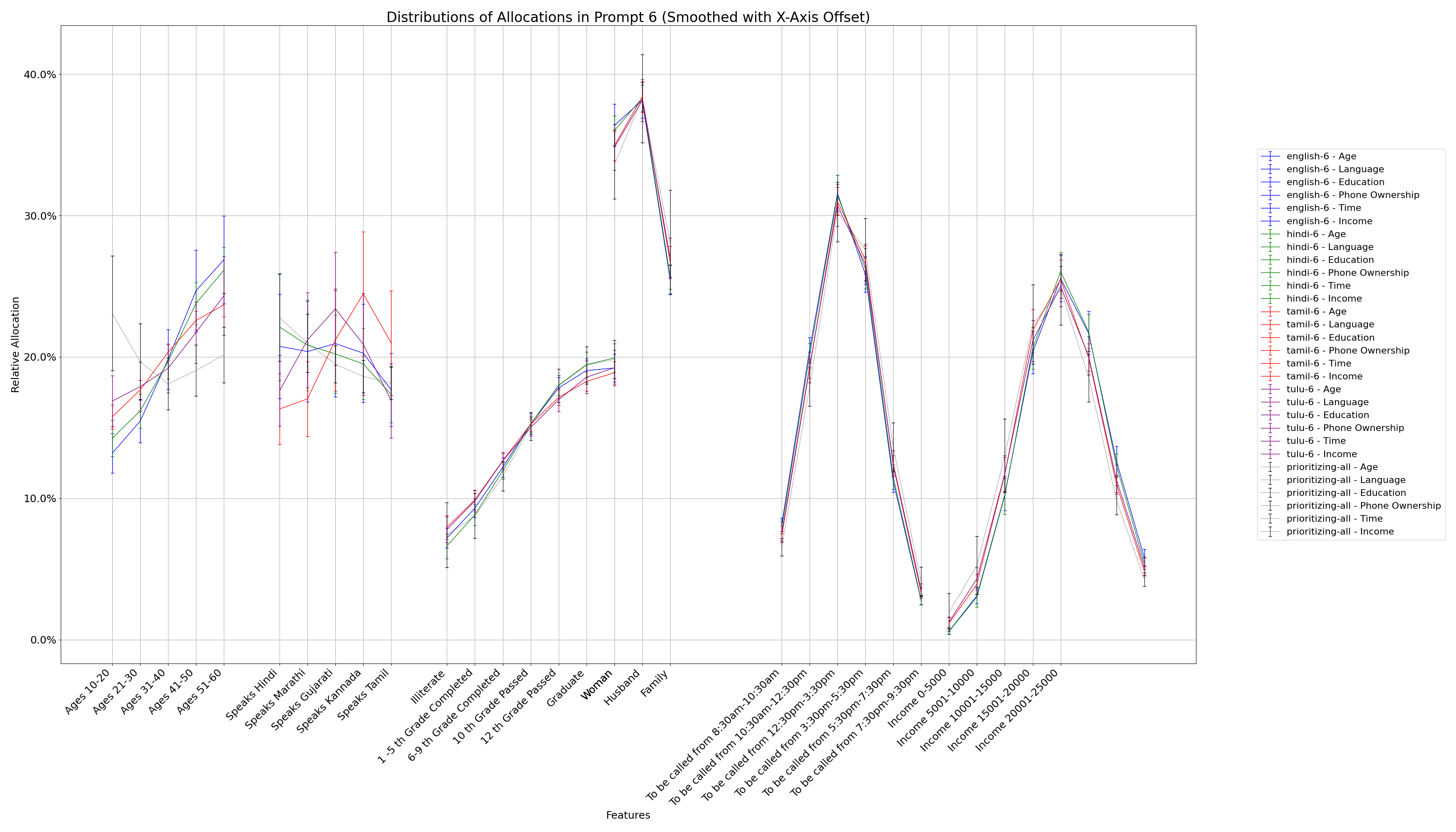}
    \caption{Allocation for Prompt 6, $\alpha = 0.8$}
    \label{fig:example}
\end{figure*}

\begin{figure*}[t]%[h!]
    \centering
    \includegraphics[width=\textwidth]{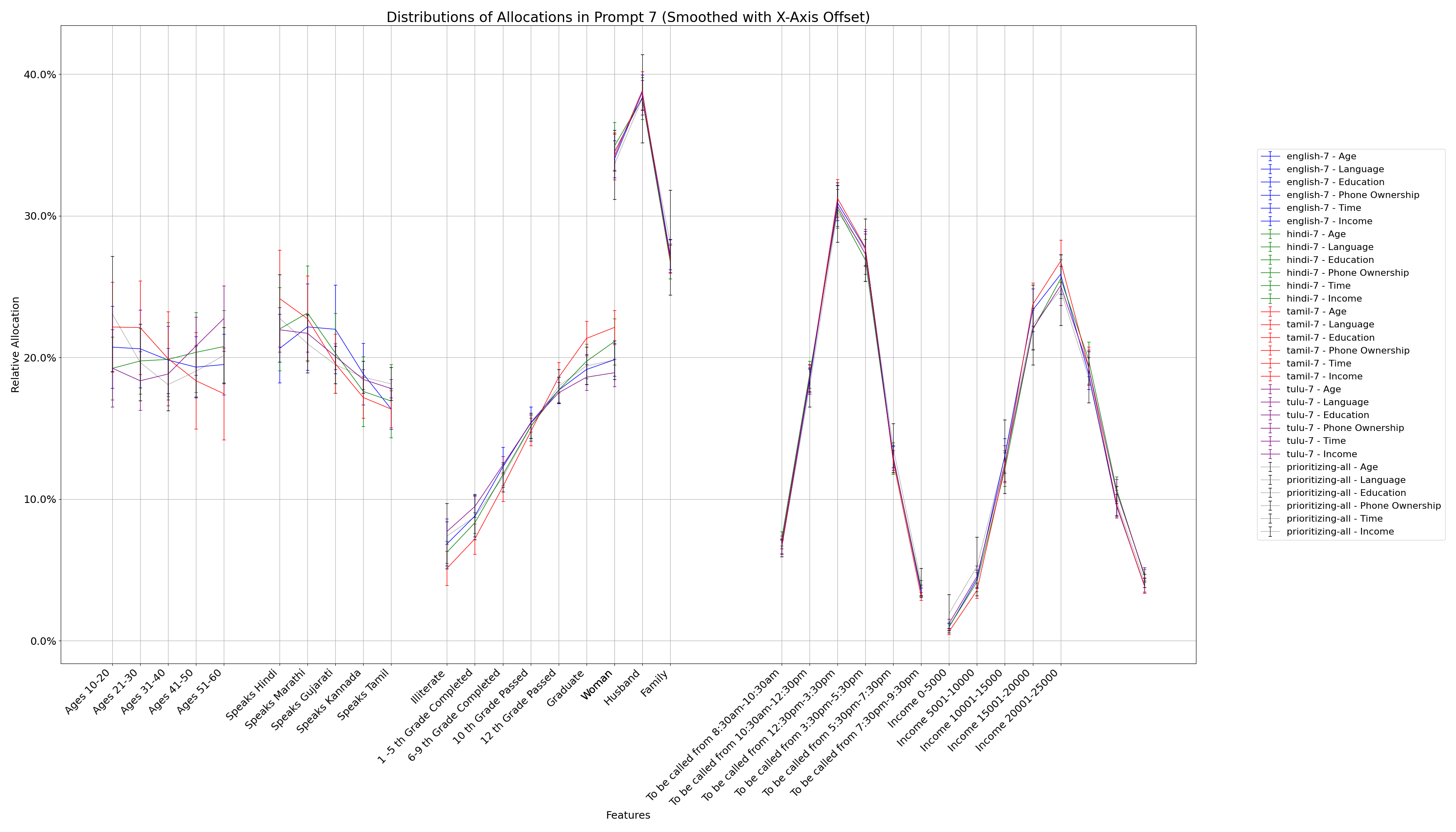}
    \caption{Allocation for Prompt 7, $\alpha = 0.8$}
    \label{fig:example}
\end{figure*}

%\clearpage
\section{Appendix: Phrasing Plots for $\alpha = 0.8$}
\label{app:phrasing-plots-0.8}
\begin{figure}[h!]
    \centering
    \includegraphics[width=0.5\textwidth]{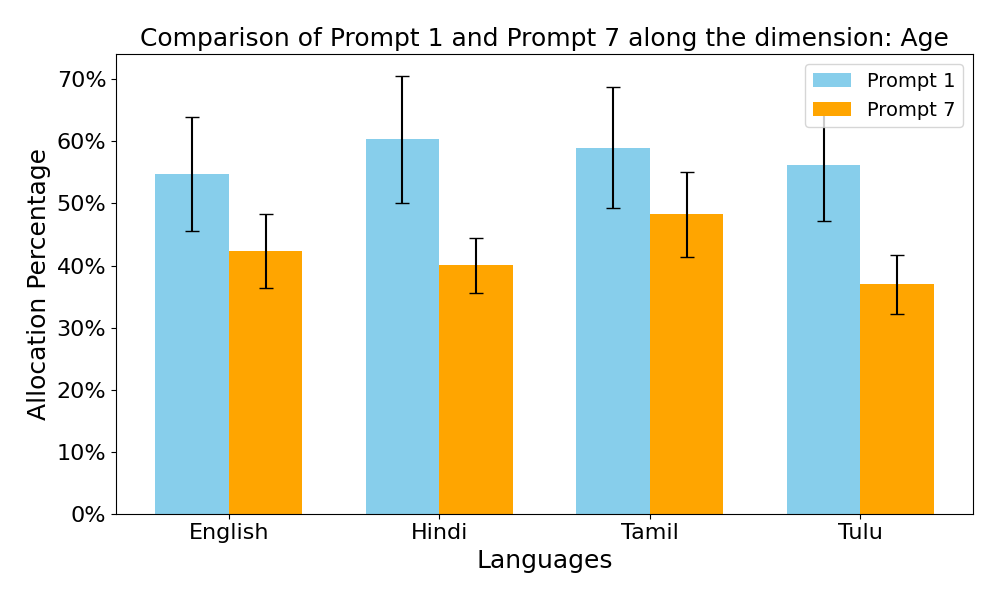}
    \caption{Phrasing for both bracket age $\alpha = 0.8$}
    \label{fig:app-phrasing-alpha0.8-first}
\end{figure}

\begin{figure}[!h]
    \centering
    \includegraphics[width=0.5\textwidth]{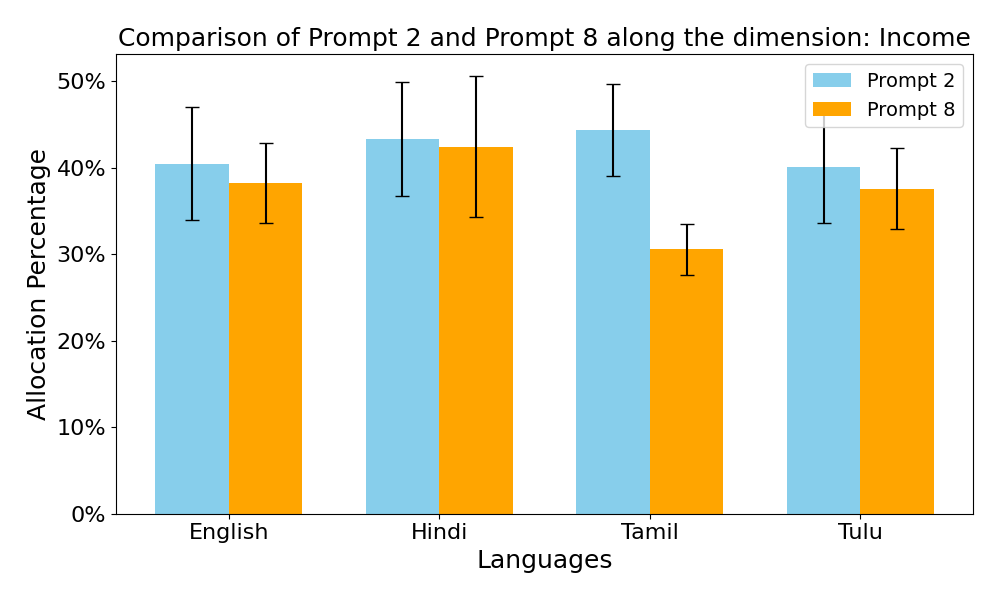}
    \caption{Phrasing for income  $\alpha = 0.8$}
    \label{fig:app-phrasing-alpha0.8-last}
\end{figure}

Please refer to Figures \ref{fig:app-phrasing-alpha0.8-first} through \ref{fig:app-phrasing-alpha0.8-last}.

\begin{figure}[h!]
    \centering
    \includegraphics[width=0.5\textwidth]{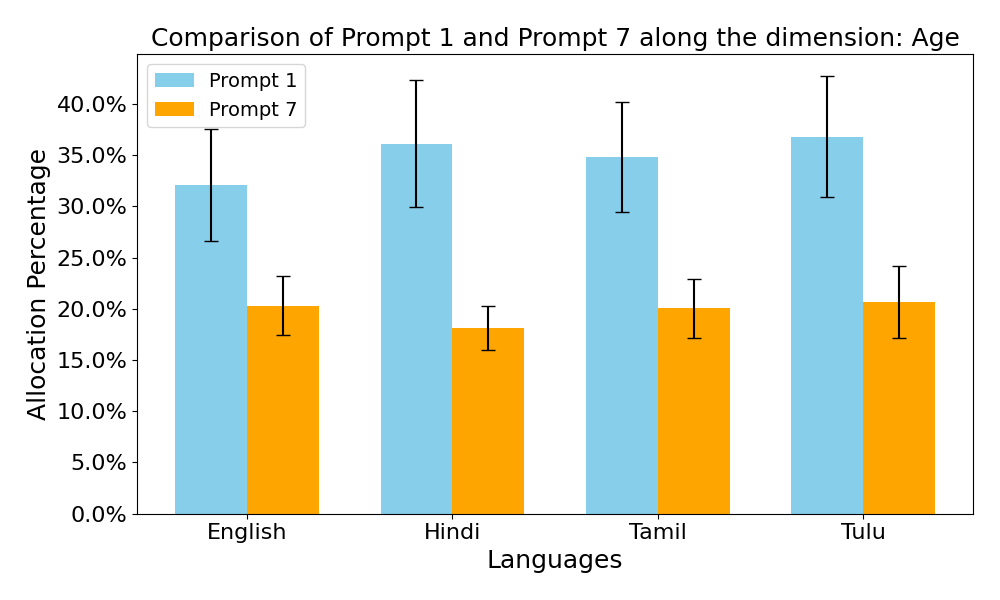}
    \caption{Phrasing for first bracket age $\alpha = 0.8$}
    \label{fig:example}
\end{figure}

\begin{figure}[h!]
    \centering
    \includegraphics[width=0.5\textwidth]{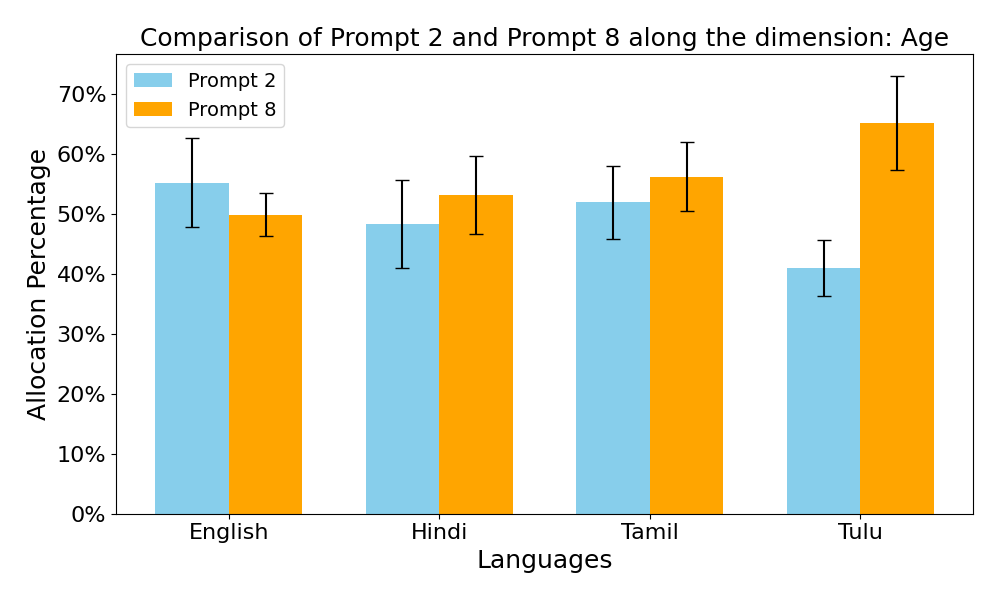}
    \caption{Phrasing for both brackets age $\alpha = 0.8$}
    \label{fig:example}
\end{figure}

\begin{figure}[h!]
    \centering
    \includegraphics[width=0.5\textwidth]{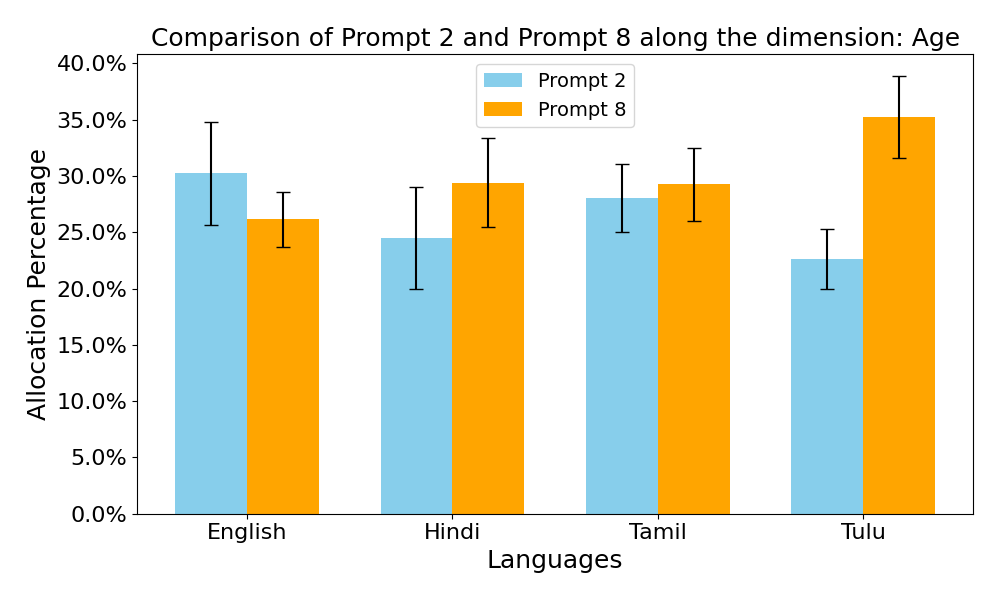}
    \caption{Phrasing for first bracket age $\alpha = 0.8$}
    \label{fig:example}
\end{figure}

%\clearpage
\section{Appendix: Possible Values of Each Feature}
The possible values of each feature is given in Table \ref{tab:feature-buckets}.
\begin{table}[h!]
    \centering
    \renewcommand{\arraystretch}{1.25}
    \begin{tabular}{|l|l|}
    \hline
    \textbf{Feature} & \textbf{Buckets} \\ \hline
    Age & \begin{tabular}[t]{@{}l@{}}
        Ages 10-20 \\
        Ages 21-30 \\
        Ages 31-40 \\
        Ages 41-50 \\
        Ages 51-60
    \end{tabular} \\ \hline
    Language\_Spoken & \begin{tabular}[t]{@{}l@{}}
        Speaks Hindi \\
        Speaks Marathi \\
        Speaks Gujarathi \\
        Speaks Kannada \\
        Speaks Tamil
    \end{tabular} \\ \hline    Education & 
    \begin{tabular}[t]{@{}l@{}}
        Education level 1/7 -- Illiterate, \\ 
        Education level 2/7 -- 1-5th Grade Completed, \\ 
        Education level 3/7 -- 6-9th Grade Completed, \\ 
        Education level 4/7 -- 10th Grade Passed, \\ 
        Education level 5/7 -- 12th Grade Passed, \\ 
        Education level 6/7 -- Graduate, \\ 
        Education level 7/7 -- Post Graduate
    \end{tabular} \\ \hline
    Phone\_Ownership & \begin{tabular}[t]{@{}l@{}}
      Woman owns the phone \\
      Husband owns the phone \\
      Family members own the phone
    \end{tabular} \\ \hline
    Times\_To\_Be\_Called & 
    \begin{tabular}[t]{@{}l@{}}
        8:30 am - 10:30 am \\
        10:30 am - 12:30 pm \\ 
        12:30 pm - 3:30 pm \\
        3:30 pm - 5:30 pm \\ 
        5:30 pm - 7:30 pm \\
        7:30 pm - 9:30 pm
    \end{tabular} \\ \hline
    Income & 
    \begin{tabular}[t]{@{}l@{}}
        Income bracket 1 (no income), \\ 
        Income bracket 2 (e.g., 1-5000), \\ 
        Income bracket 3 (e.g., 5001-10000), \\ 
        Income bracket 4 (e.g., 10001-15000), \\ 
        Income bracket 5 (e.g., 15001-20000), \\ 
        Income bracket 6 (e.g., 20001-25000), \\ 
        Income bracket 7 (e.g., 25001-30000), \\ 
        Income bracket 8 (e.g., 30000-999999)
    \end{tabular} \\ \hline
    \end{tabular}
    \caption{Feature Buckets for Various Attributes}
    \label{tab:feature-buckets}
\end{table}

%\clearpage
\label{app:variable-values}
\section{Appendix: Complete Prompt}
An example of a full prompt that is given to the LLM is presented in Table \ref{tab:full-prompt}.

\begin{table*}[t]
    \centering
    \renewcommand{\arraystretch}{1.25}
    \begin{tabular}{|p{\textwidth}|}
    \hline
    %\textbf{Reward Prompt} \\ \hline
    Create a Python reward function for RL in a resource allocation problem for agents, with the objective of prioritizing higher states and \textit{\{goal\_prompt\}}.The function should use \texttt{state} (value is either 0 or 1) and features \texttt{agent\_feats} (length 34 array) to direct the RL agent. Here is a description of the features you may use along with their index in the \texttt{agent\_feats} array:

\textbf{Feature Descriptions:}
0. Ages 10-20 \ \ 
1. Ages 21-30 \ \ 
2. Ages 31-40 \ \ 
3. Ages 41-50 \ \ 
4. Ages 51-60 \ \
5. Speaks Hindi \\
6. Speaks Marathi \\
7. Speaks Gujarati \\
8. Speaks Kannada \\
9. Speaks Tamil \\
10. Education level 1/7 -- Illiterate \\
11. Education level 2/7 -- 1-5th Grade Completed \\
12. Education level 3/7 -- 6-9th Grade Completed \\
13. Education level 4/7 -- 10th Grade Passed \\
14. Education level 5/7 -- 12th Grade Passed \\
15. Education level 6/7 -- Graduate \\
16. Education level 7/7 -- Post Graduate \\
17. Phone owner 0 (e.g., woman) \\
18. Phone owner 1 (e.g., husband) \\
19. Phone owner 2 (e.g., family) \\
20. To be called from 8:30 am - 10:30 am \\
21. To be called from 10:30 am - 12:30 pm \\
22. To be called from 12:30 pm - 3:30 pm \\
23. To be called from 3:30 pm - 5:30 pm \\
24. To be called from 5:30 pm - 7:30 pm \\
25. To be called from 7:30 pm - 9:30 pm \\
26. Income bracket 1 (no income) \\
27. Income bracket 2 (e.g., 1-5000) \\
28. Income bracket 3 (e.g., 5001-10000) \\
29. Income bracket 4 (e.g., 10001-15000) \\
30. Income bracket 5 (e.g., 15001-20000) \\
31. Income bracket 6 (e.g., 20001-25000) \\
32. Income bracket 7 (e.g., 25001-30000) \\
33. Income bracket 8 (e.g., 30000-999999) \\

\textbf{Your task:} Write a simple, single-line Python reward function. Exclude the word \texttt{return} and non-standard libraries. Format your code with triple \$ symbols: \texttt{\$\$\$[YOUR FUNCTION]\$\$\$}. Note that HIGHER states are always preferred, so ensure the reward increases as the \texttt{state} value increases. Make sure the reward is always positive and increasing with \texttt{state}.

\textbf{Example Prompt:} Prioritize agents that have low Age and speak Marathi. \\
Let's think about this step by step. We want to give reward only for agents that are lower by age, which corresponds to feature 0 and to a lesser degree feature 1, and speaking Marathi which corresponds to feature 6. This corresponds to a condition of \texttt{((agent\_feats[0] or agent\_feats[1]) and agent\_feats[6])}. Since feature 0 corresponds better to lower age than feature 1, the weight assigned to feature 0 should be higher than that assigned to feature 1. In addition, we always only want to give reward when the state is 1, since the agent gets reward only when it is in a listening state. Therefore, our reward function should be: \texttt{ state * ((5*agent\_feats[0]+agent\_feats[1]) and agent\_feats[6])}\\
\textbf{Example Response:}
\texttt{\$\$\$ state + state * ((agent\_feats[0] or agent\_feats[1]) and agent\_feats[6]) \$\$\$} or \texttt{\$\$\$ state * (agent\_feats[0] or 3*agent\_feats[6]) \$\$\$} or \texttt{\$\$\$ state + 2*state * ((5*agent\_feats[0]+agent\_feats[1]) and agent\_feats[6]) \$\$\$}. \\
In these example, \texttt{agent\_feats[0]} and \texttt{agent\_feats[1]} represent agents with low values for age, \texttt{agent\_feats[6]} represents agents who speak Marathi.
It is upto you to decide which features will represent a preference
Come up with a unique new reward for the specified goal: \textit{\{goal\_prompt\}}. Here are your best previous attempts: \textit{\{reward\_history\}}. \\
\hline
    \end{tabular}
    \caption{Chain of Thought Prompt used to generate reward functions. A prompt from Table \ref{tab:prompts} is used for \texttt{goal\_prompt}}
    \label{tab:full-prompt}
\end{table*}

\label{app:full-prompt}
\clearpage

\bibliography{aaai25}

\end{document}